\documentclass[10pt,twocolumn,letterpaper]{article}
\usepackage[pagenumbers]{cvpr}
\usepackage{graphicx}
\usepackage{amsmath,amssymb}
\usepackage{booktabs}
\usepackage{multirow}
\usepackage{tabularx}
\usepackage{xcolor}
\usepackage{microtype}

\usepackage[most]{tcolorbox}
\definecolor{promptblue}{HTML}{D6E6FF}
\definecolor{promptcyan}{HTML}{D7F9F8}
\definecolor{promptyellow}{HTML}{FFFFEA}
\definecolor{promptorange}{HTML}{FFF0D4}
\definecolor{promptred}{HTML}{FBE0E0}
\definecolor{promptlavender}{HTML}{E5D4EF}
\definecolor{promptlilac}{HTML}{F2EAF7}
\definecolor{promptcream}{HTML}{F9F5FB}

\definecolor{cvprblue}{rgb}{0.21,0.49,0.74}
\usepackage[pagebackref,breaklinks,colorlinks,allcolors=cvprblue]{hyperref}
\def\confName{arXiv preprint}
\def\confYear{2026}
\title{PhysQ2QBench: Trustworthy Evaluation of Physical Consistency in Video Generation from Qualitative to Quantitative}
\author{
Cui Yakun$^{1,*}$ \quad
Bocheng Zeng$^{2,3,*}$ \quad
Wangchenhui Wu$^{2,3,*}$ \quad
Hang Du$^{4}$ \\
Baijia Yu$^{2,5}$ \quad
Tiantian Geng$^{6}$ \quad
Chenye Wang$^{1}$ \quad
Yuyao Zhang$^{1}$ \quad
Sirui Han$^{1,\dagger}$ \\[0.4em]
{\small $^{1}$The Hong Kong University of Science and Technology \quad
$^{2}$Shenlan Lab} \\
{\small $^{3}$Beijing University of Posts and Telecommunications \quad
$^{4}$Zhongguancun Laboratory} \\
{\small $^{5}$University of Electronic Science and Technology of China \quad
$^{6}$Southern University of Science and Technology} \\[0.25em]
{\small Correspondence: \texttt{ycuibd@connect.ust.hk}} \\
{\small $^{*}$Equal contribution. $^{\dagger}$Corresponding author.}
}

\makeatletter
\let\physqOriginalMaketitle\@maketitle
\def\@maketitle{%
  \physqOriginalMaketitle
  \begin{center}
  \includegraphics[width=0.85\textwidth]{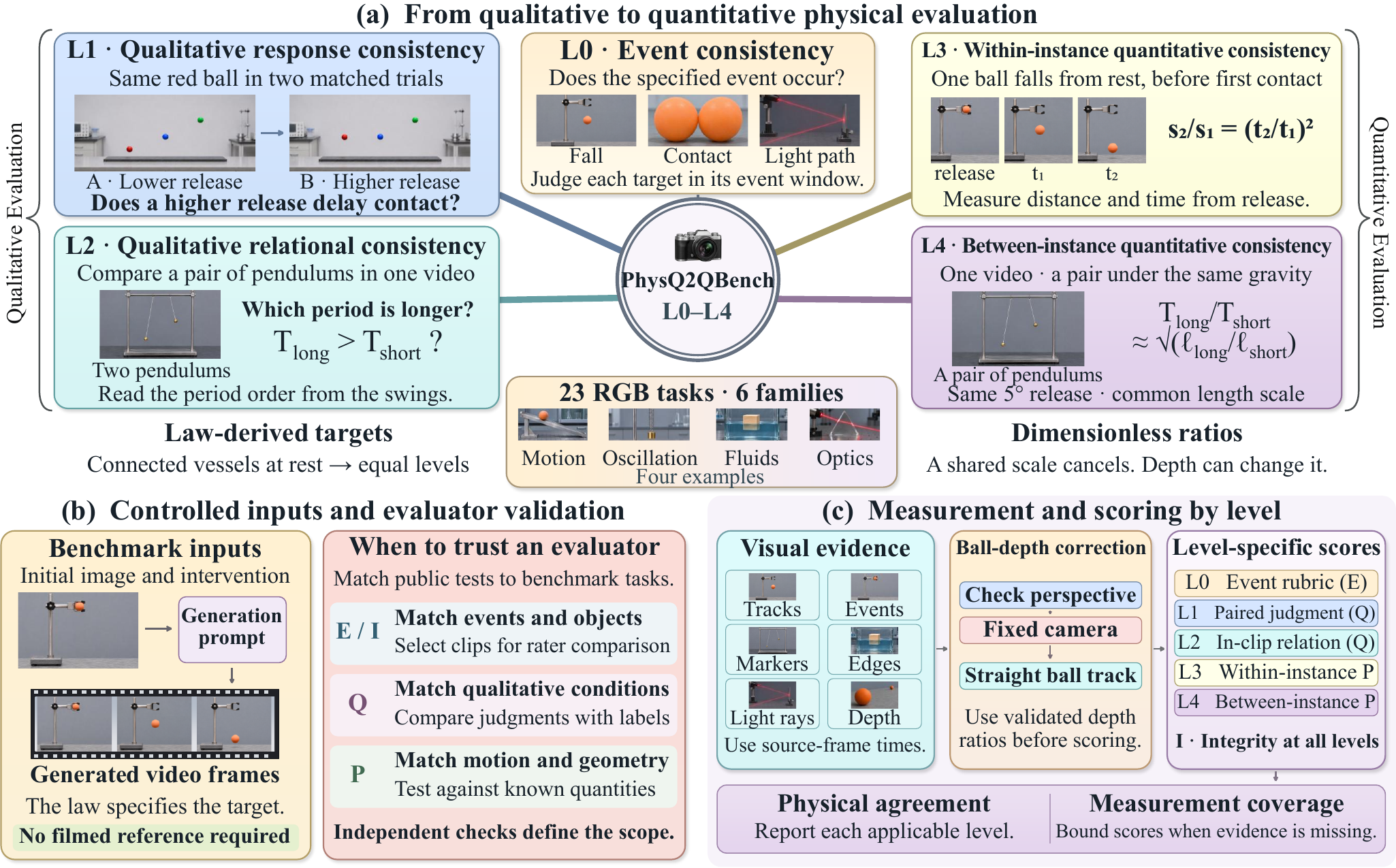}
  \captionsetup{hypcap=false}
  \captionof{figure}{PhysQ2QBench tests whether the expected event occurs (L0), whether responses to interventions (L1) and relations between instances (L2) follow qualitative predictions, and whether measurements within one instance (L3) or between instances (L4) agree with a physical law. The law sets the target. For example, the same liquid in communicating vessels at rest reaches equal heights under equal surface pressure. Dimensionless ratios cancel a shared scale, but motion in depth can change it (\cref{fig:projection}). We validate each evaluator on examples matched to its events, relations or viewing conditions.}
  \label{fig:framework}
  \end{center}
}
\makeatother
\begin{document}
\raggedbottom
\maketitle
\begin{abstract}
Video generators serve as world models and supply training data for embodied agents. A falling ball can look convincing even when it accelerates at the wrong rate. When qualitative and quantitative tests use different scenes, it is hard to separate errors in the physics from differences in scene difficulty.
\textbf{PhysQ2QBench} tests physical consistency at five levels on the same controlled scenes, from qualitative behavior to numerical agreement. Its 594 inputs support 1,231 evaluation samples across 23 tasks in six physics families, including optics, electromagnetism and thermal physics, with a separate acoustic extension. We validate evaluators on matched examples, checking qualitative judgments against public annotations or simulator states and depth estimates against independent 3D trajectories. We derive dimensionless targets from physical laws without filmed references. Ratios cancel a common image scale, but motion in depth can change that scale across frames. Our depth correction reduces depth-ratio error from 12.5\% to 4.6\% on motion-capture data. In generated videos, correction changes quantitative scores more on tracks with larger depth excursions.
The two proprietary models lead the six open-weight models on all three qualitative levels (mean L0 scores of 3.41 versus 2.53 out of 5). Between-instance scores fall below within-instance scores in 42 of 48 model--family comparisons. For MiniMax-H3 and Seedance 2.5, mean agreement rises from the early to the middle window for each of four laws. Six of the eight curves then decline, including all four for uniform motion and constant acceleration. Objects in the same scene respond more similarly than the law predicts in 77.2\% of MiniMax-H3 videos and 73.7\% of Seedance 2.5 videos. MiniMax-H3 retains less of the predicted contrast (median 0.168 versus 0.330).
\end{abstract}

\section{Introduction}
\label{sec:intro}
Video generators can create and restyle footage~\cite{kondratyuk2024videopoet,bartal2024lumiere} or edit a source video from a text prompt~\cite{wu2023tuneavideo}. As world models, they predict the consequences of actions~\cite{yang2024unisim,bruce2024genie} and can simulate playable environments~\cite{valevski2025gamengen}. In robotics, generated videos serve as visual plans from which an agent can infer actions~\cite{du2023unipi} or as demonstrations for training policies~\cite{jang2025dreamgen}. VBench~\cite{huang2024vbench} and EvalCrafter~\cite{liu2024evalcrafter} assess visual consistency and motion quality. A falling ball can look consistent and move smoothly yet reach the ground too soon. For a robot learning when to catch it, that timing error matters.

VideoPhy~\cite{bansal2024videophy} evaluates interactions between materials, and VideoPhy-2~\cite{videophy2} examines the physical rules governing actions. PhyGenBench~\cite{phygenbench} and T2VPhysBench~\cite{guo2025t2vphysbench} test whether events follow specified laws, while PhyGround~\cite{phyground} provides human judgments of law adherence and object persistence. Showing that a released ball falls is a first step. In free fall from rest, doubling the time since release should quadruple the distance fallen. Prior work compares generated motion with simulations~\cite{kang2025physical} or filmed experiments~\cite{motamed2026physicsiq}. Morpheus~\cite{morpheus} and PVBench~\cite{pvbench} measure physical quantities from generated motion, and Principia~\cite{principia} tests relations between objects in a shared scene. Across separate suites, a lower quantitative score could reflect a harder scene, a missing event, or an error in its timing or magnitude.

\begin{figure}[t]
\centering
\includegraphics[width=\columnwidth]{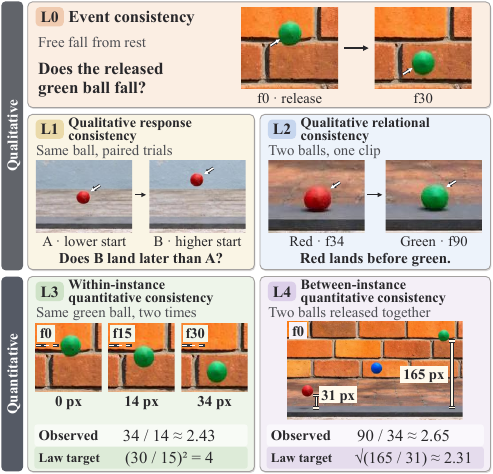}
\caption{Free fall illustrates the five criteria in Appendix Table~\ref{tab:levels}. L1 shows the initial frames of two matched trials. All other panels use clip \texttt{T01\_reference\_P4}. Time is measured from release. The laws $s(t)=gt^2/2$ and $T=\sqrt{2h/g}$ give the displacement and elapsed-time ratios tested at L3 and L4, respectively. At L4, $h$ is the distance fallen to the last descending state; contact is not required.}
\label{fig:fall-levels}
\end{figure}

PhysQ2QBench tests this progression on one collection of controlled scenes (\cref{fig:framework,fig:fall-levels}). Consider balls released from different heights. Event consistency (L0) asks whether a released ball falls. Qualitative response consistency (L1) asks whether raising the release height delays contact across matched trials. Qualitative relational consistency (L2) asks whether the lower of two balls released together lands first. L3 and L4 ask how closely measured quantities follow the law. Within-instance quantitative consistency (L3) relates one ball's displacement to elapsed time. Between-instance quantitative consistency (L4) relates elapsed fall times to the corresponding distances fallen, using each ball's last descending state without requiring contact. The same distinction applies to other physical laws, from liquid levels in connected vessels to the angles of reflected and refracted rays.

\begin{table*}[t]
\begin{minipage}{\textwidth}
\centering
\begingroup
\small
\setlength{\tabcolsep}{0.9pt}
\renewcommand{\arraystretch}{1.00}
\newcommand{\pqh}[2]{\shortstack{\bfseries #1\\\bfseries #2}}
\newcommand{\pqY}{\textcolor[rgb]{0.00,0.43,0.23}{$\checkmark$}}
\newcommand{\pqP}{\textcolor[rgb]{0.64,0.40,0.04}{$\triangle$}}
\newcommand{\pqN}{\textcolor[rgb]{0.68,0.20,0.21}{$\times$}}
\begin{tabular*}{\linewidth}{@{\extracolsep{\fill}}l*{13}{c}@{}}
\toprule
& \multicolumn{4}{c}{\textbf{Physical targets}}
& \multicolumn{4}{c}{\textbf{Qualitative to quantitative}}
& \multicolumn{5}{c}{\textbf{Measurement reliability}} \\
\cmidrule(lr){2-5}\cmidrule(lr){6-9}\cmidrule(l){10-14}
\textbf{Benchmark} & \pqh{Multi-}{domain} & \pqh{Law}{target} & \pqh{Film-}{free} & \pqh{Unit-}{free} & \pqh{Graded}{levels} & \pqh{Paired}{trials} & \pqh{Object}{pairs} & \pqh{Joint}{score} & \pqh{Meas.}{audit} & \pqh{Depth}{corr.} & \pqh{Matched}{tests} & \pqh{Meas.}{coverage} & \pqh{Score}{bounds} \\
\midrule
PhyGenBench~\cite{phygenbench} & \pqY & \pqN & \pqY & \pqN & \pqN & \pqN & \pqN & \pqN & \pqN & \pqN & \pqP & \pqN & \pqN \\
VideoPhy-2~\cite{videophy2} & \pqY & \pqN & \pqY & \pqN & \pqN & \pqN & \pqN & \pqN & \pqN & \pqN & \pqP & \pqN & \pqN \\
Physics-IQ Verified~\cite{physicsiqverified} & \pqY & \pqN & \pqN & \pqN & \pqN & \pqN & \pqN & \pqN & \pqY & \pqN & \pqP & \pqN & \pqN \\
Morpheus~\cite{morpheus} & \pqN & \pqY & \pqP & \pqY & \pqN & \pqN & \pqN & \pqP & \pqY & \pqN & \pqP & \pqP & \pqN \\
Principia~\cite{principia} & \pqN & \pqY & \pqP & \pqY & \pqP & \pqN & \pqY & \pqP & \pqY & \pqN & \pqP & \pqN & \pqN \\
World Models' Last Exam~\cite{lastExam} & \pqY & \pqY & \pqY & \pqY & \pqP & \pqN & \pqY & \pqP & \pqP & \pqN & \pqP & \pqP & \pqN \\
PVBench~\cite{pvbench} & \pqY & \pqY & \pqY & \pqY & \pqP & \pqN & \pqY & \pqP & \pqP & \pqP & \pqP & \pqY & \pqN \\
\midrule
\textbf{PhysQ2QBench (ours)} & \pqY & \pqY & \pqY & \pqY & \pqY & \pqY & \pqY & \pqY & \pqY & \pqY\rlap{$^{*}$} & \pqY & \pqY & \pqY \\
\bottomrule
\end{tabular*}
\par\smallskip
{\footnotesize \pqY\ Supported\qquad \pqP\ Partial support\qquad \pqN\ Unreported in the cited protocol}
\endgroup
\captionsetup{hypcap=false,justification=raggedright,singlelinecheck=false}
\captionof{table}{Physical-consistency benchmarks. A law supplies a numerical target. Film-free covers both generation and scoring ($\triangle$ for filmed input only). Paired trials vary a condition across clips. Object pairs share a clip. Joint scores combine qualitative and numerical evidence for the same comparison. Measurement audits test sensitivity to imaging or extraction. Matched tests select independent data for the evaluator's task. Coverage records the share measured, and score bounds retain the weight of unresolved comparisons. $^{*}$We detect and correct for changes in depth that affect measurements.}
\label{tab:benchmark-comparison}
\end{minipage}

\end{table*}

Interpreting a score at any level requires knowing what serves as its reference. A human or a vision--language model can judge an event directly, but its reliability depends on agreement with independent judgments under comparable conditions. Physics-IQ compares a generated continuation with a filmed outcome~\cite{motamed2026physicsiq}. Camera pose, calibration, and release conditions can change the observed trajectory even when both trials obey the law. Physics-IQ Verified found that revising prompts, filmed references, and scoring weights changed generator rankings~\cite{physicsiqverified}. Table~\ref{tab:benchmark-comparison} compares the targets and measurement checks used by these benchmarks.

We match event and object-persistence tests to human-rated PhyGround videos~\cite{phyground} by task description and first-frame appearance. For qualitative relations, we match each event window to a PhysBench annotation~\cite{physbench} or a PhysInOne simulator state~\cite{physinone}. On these matched videos, GPT-6.1-sol reproduces the reference qualitative score exactly in 95.3\% of cases (\cref{sec:experiments}).

At L3 and L4, physical laws supply the target without a filmed reference. Connected vessels holding the same liquid at rest under equal surface pressure must reach equal levels. A displacement ratio cancels a shared image scale, but motion in depth can make a true ratio of 4 appear as 3.67 even with a fixed camera (\cref{fig:projection}).

We detect and correct changes in depth when they affect a measurement. For straight ball tracks in a fixed view, our evaluator estimates relative depth from Depth Anything~3 features~\cite{da3} and uses apparent width to guide correction. On ABCD motion capture~\cite{abcd}, mean depth-ratio error is 4.6\%, versus 12.5\% for a fixed-depth baseline (\cref{fig:depth-methods}). Each score includes the fraction of comparisons we could measure.

The benchmark registers 1{,}231 evaluation samples on 594 inputs across 23 tasks. Alongside motion and energy, the RGB tasks cover oscillation and statics, fluids, optics and geometry, electromagnetism, and thermal physics. A separate acoustic extension covers echoes.

Across eight generators, the two proprietary models lead the six open-weight models on all three qualitative levels (\cref{sec:main-results}). The respective group means are 3.41 and 2.53 out of 5 at L0, 0.69 and 0.31 at L1, and 0.67 and 0.33 at L2. The proprietary models also reproduce some numerical relations, averaging 0.76 at L3 and 0.67 at L4. Across all models, L4 scores fall below L3 scores in 42 of 48 model--family comparisons. Optics and geometry has the lowest mean L1 score (0.12). Models often fail to change the reflected or refracted ray appropriately when the incident ray changes.

On generated trajectories, the larger the change in depth, the more perspective correction alters the L3 and L4 scores (\cref{fig:generator-depth-effect}a). Within a clip, physical agreement also varies with time. MiniMax-H3 and Seedance 2.5 score higher in the middle window than in the early window for each of the four laws tested. By the late window, six of the eight model--law curves have fallen from their midpoint values, including all four for uniform motion and constant acceleration (\cref{fig:generator-depth-effect}b). The two models also understate differences between objects in the same scene. MiniMax-H3 and Seedance 2.5 underestimate contrast in 77.2\% and 73.7\% of videos, with median contrast retention of 0.168 and 0.330, respectively (\cref{fig:response-contrast}).

\section{Related Work}
\label{sec:related-work}

PhyGenBench and VideoPhy-2 use human and model judgments to assess physical commonsense~\cite{phygenbench,videophy2}. Physics-IQ Verified reports a change in model rankings after revising prompts, filmed references, and example weights~\cite{physicsiqverified}. Morpheus tests conservation laws~\cite{morpheus}, while Principia compares objects governed by the same law~\cite{principia}. World Models' Last Exam in Physics and PVBench measure physical behavior across domains~\cite{lastExam,pvbench}. WorldBench separates intuitive judgments from estimates of physical parameters such as friction coefficients~\cite{worldbench}. PhysQ2QBench scores event occurrence, responses to interventions and relations between instances separately from numerical agreement within and between instances (Appendix Table~\ref{tab:levels}).

Measurement errors can mimic physical violations. GAUGE uses calibrated experiments~\cite{gauge}, and WorldBench and Last Exam validate measurements under controlled conditions~\cite{worldbench,lastExam}. PDI-Bench checks scale--depth consistency, motion, and rigidity~\cite{pdibench}. RigidBench tests metrics with controlled perturbations and aligns predicted disparity to reference disparity~\cite{rigidbench}. Our validation sets match the event, relation or geometry each evaluator must recover. For depth, we check the correction against independent 3D measurements before comparing dimensionless ratios with the law. The physical prediction and a filmed reference trajectory play no part in fitting the correction.

Metrology formalizes uncertainty propagation~\cite{jcgm2008gum}, and TaskUQ calibrates intervals for quantities inferred from reconstructed images~\cite{taskuq}. Our optical tests check interval coverage on held-out measurements. Joint Physical Fidelity retains the weights of confirmed failures and bounds the contributions of unresolved comparisons.

\begin{figure*}[!t]
\centering
\includegraphics[width=\textwidth]{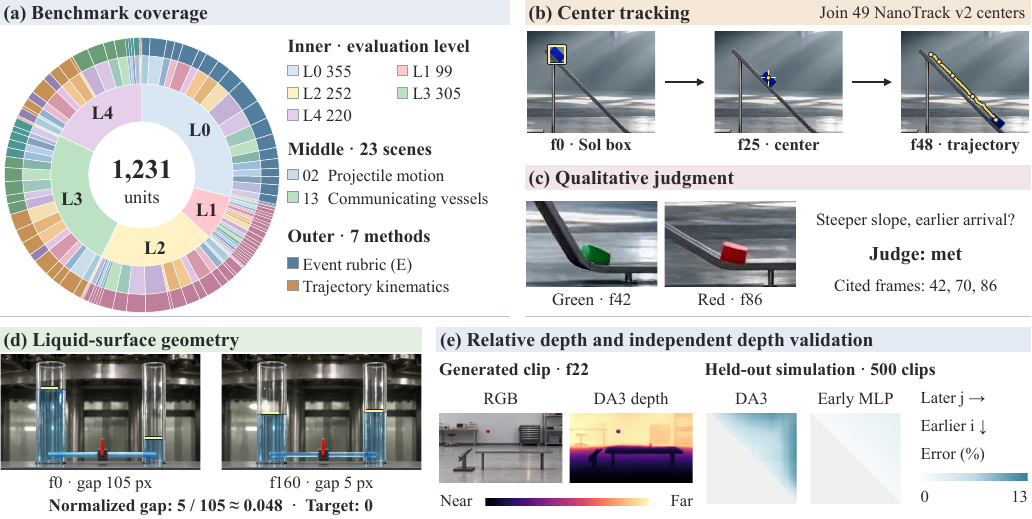}
\caption{(a) The 1,231 evaluation samples by level, scene and method (full key in Appendix Fig.~\ref{fig:inventory-key}). Each L1 pair counts once. (b--d) Slider tracking, arrival-order evidence and liquid-level measurement. (e) DA3 scene depth and frame-pair depth-ratio errors on held-out simulations.}
\label{fig:inventory}
\end{figure*}

\section{From Events to Quantitative Laws}
\label{sec:benchmark}

\subsection{Five levels of physical consistency}
Appendix Table~\ref{tab:levels} defines the five levels of physical consistency. An instance can be an object, a light beam or an apparatus. In \cref{fig:fall-levels}, a released ball falls (L0). Raising its release point should delay contact in a matched trial (L1), and the lower of two balls released together should land first (L2). L3 tests one ball's displacement over time. L4 tests the ratio of the two fall times against their release heights. We assign each test to a level before generation and score the levels independently.

Table~\ref{tab:relation-examples} shows how the same law gives different tests at L3 and L4. One pendulum should keep its period across cycles. For two pendulums, their lengths predict the ratio of their periods. Appendix~\ref{app:relation-catalog}, Table~\ref{tab:relation-catalog}, gives the remaining relations and their measurement conditions.

\begin{table}[t]
\centering
\caption{Physical relations within one instance (L3) and between instances in the same scene (L4). Measurement conditions and notation are in Appendix~\ref{app:relation-catalog}.}
\label{tab:relation-examples}
\begingroup
\small
\setlength{\tabcolsep}{2pt}
\renewcommand{\arraystretch}{1.15}
\begin{tabular*}{\columnwidth}{@{\extracolsep{\fill}}lcc@{}}
\toprule
\textbf{Scene} & \textbf{L3} & \textbf{L4} \\
\midrule
Free fall & $\dfrac{h_2}{h_1}=\!\left(\dfrac{t_2}{t_1}\right)^{\!2}$ & $\dfrac{t_i}{t_j}=\sqrt{h_i/h_j}$ \\
\addlinespace[3pt]
Pendulum & $\dfrac{T_2}{T_1}\approx 1$ & $\dfrac{T_i}{T_j}\approx\sqrt{\ell_i/\ell_j}$ \\
\addlinespace[3pt]
\shortstack[l]{Communicating\\vessels} & $h_k^f=\dfrac{\sum_m A_m h_m^0}{\sum_m A_m}$ & $-\dfrac{\Delta h_i}{\Delta h_j}=\dfrac{A_j}{A_i}$ \\
\addlinespace[3pt]
Rod shadows & $\dfrac{s_i}{d_i}=\dfrac{H_i}{H_L-H_i}$ & $\dfrac{s_i}{s_j}=\dfrac{H_i}{H_j}$ \\
\addlinespace[3pt]
Magnet and coil & $J_{\rm out}=-J_{\rm in}$ & $\dfrac{|\bar U_i|}{|\bar U_j|}=\dfrac{\Delta t_j}{\Delta t_i}$ \\
\bottomrule
\end{tabular*}
\endgroup
\end{table}

\subsection{Benchmark coverage}
The benchmark contains 594 inputs spanning 169 apparatus and intervention configurations across 23 tasks in six physics families: motion and energy, oscillation and statics, fluids, optics and geometry, electromagnetism, and thermal physics. The inputs reflect the available background variants for each configuration. The inputs support 1,231 evaluation samples: 355 at L0, 99 at L1, 252 at L2, 305 at L3 and 220 at L4 (\cref{fig:inventory}). Each L1 sample is a controlled pair; the 99 pairs involve 183 inputs. Data construction and a generation prompt are provided in Appendix~\ref{app:construction-evaluation}.

\setcounter{dbltopnumber}{1}
\section{Trustworthy Physical Measurement}
\label{sec:method}

\subsection{Choosing a physical reference}
Each quantitative task specifies a law, the conditions under which it holds and the observations needed to test it. In a communicating vessel, equality of free-surface heights requires hydraulic connection, the same liquid, equal gas pressure above the limbs and sufficient time to equilibrate. The evaluator locates each tube and detects the air--liquid boundary in a narrow vertical strip (\cref{fig:inventory}). We measure the two heights against a common vertical datum and divide their residual difference by the initial height span when it is resolvable. At equilibrium this normalized difference should be zero. In a free-fall test, the expected height ratio comes from the elapsed times. The generator is free to produce any appearance or trajectory consistent with the specified conditions.

\begin{figure}[t]
\centering
\includegraphics[width=\columnwidth]{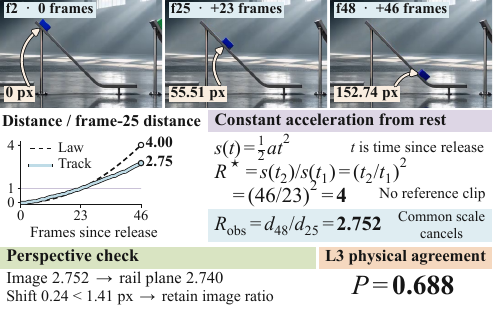}
\caption{L3 test on benchmark clip \texttt{T04\_E01\_P2}. Frame labels give along-rail displacement from release. The measured ratio falls $31.2\%$ below the square-law prediction in \cref{eq:fall}.}
\label{fig:l3-incline-example}
\end{figure}

For motion from rest with constant acceleration, $s(t)=at^2/2$ with time measured from release. Free fall gives $a=g$. On a straight incline of fixed angle $\theta$, the vertical drop is $s\sin\theta$ and the same distance ratio applies while the acceleration remains constant. For two pre-contact observations $i$ and $j$, the predicted ratio is
\begin{equation}
R^\star_{ji}=\frac{s(t_j)}{s(t_i)}=\left(\frac{t_j}{t_i}\right)^2 .
\label{eq:fall}
\end{equation}
At constant frame rate, elapsed frame counts from release give the time ratio directly. For nonuniform sampling we use the actual presentation times. We exclude an incline's curved runout from the straight-segment test. We fix the time window and identify the objects before measuring displacement. In \cref{fig:l3-incline-example}, GPT-6.1 Sol locates the blue slider in the first frame. We join its successive NanoTrack~v2 box centers to trace its path along the straight section (\cref{fig:inventory}). To check how much perspective affects its measured displacement, we reconstruct the rail plane from its two fixed supports and DA3-estimated camera intrinsics, independently of the square-law prediction.

\begin{figure*}[t]
\centering
\includegraphics[width=\textwidth]{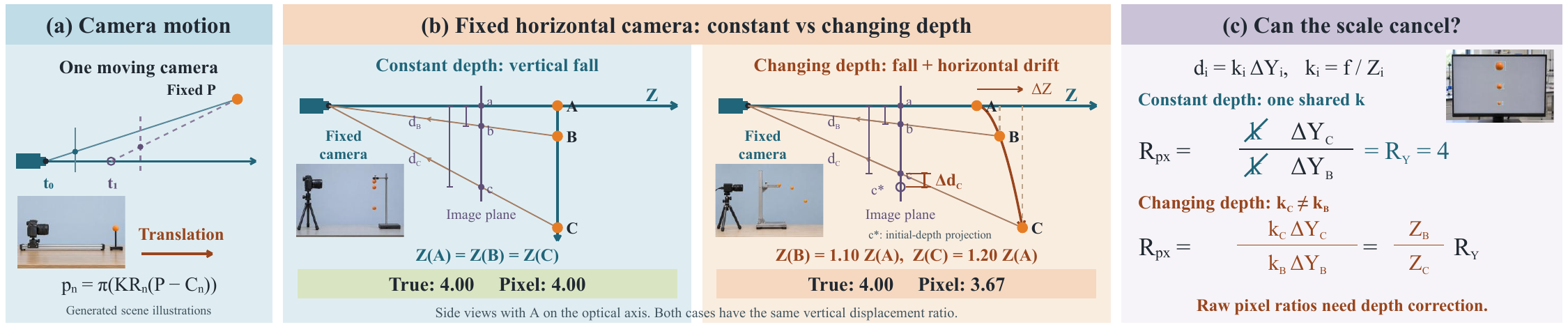}
\caption{Perspective can change a dimensionless ratio. With the release point on the optical axis of a fixed horizontal camera, the same vertical-displacement ratio of 4 projects to 4 at constant depth and to $4(1.10/1.20)\simeq 3.67$ at the indicated depths. The drifting path illustrates the projection geometry. Insets are scene illustrations.}
\label{fig:projection}
\end{figure*}

\subsection{When does image scale cancel?}
Let $\Delta Y_i$ be the vertical displacement from release and $Z_i$ the camera-space depth in the geometry of \cref{fig:projection}. A pinhole camera with focal length $f_c$ gives displacement $d_i=f_c\Delta Y_i/Z_i$. The observed ratio and its correction are
\begin{equation}
R^{\mathrm{px}}_{ji}=\frac{d_j}{d_i}
=\frac{Z_i}{Z_j}\frac{\Delta Y_j}{\Delta Y_i},\qquad
\widehat R^{\mathrm{corr}}_{ji}=R^{\mathrm{px}}_{ji}\frac{\widehat Z_j}{\widehat Z_i}.
\label{eq:correction}
\end{equation}
An error in the estimated depth ratio multiplies the corrected physical ratio, so we validate relative depth independently of the law. For an earlier fixed version of the depth estimator, each cell in \cref{fig:inventory} shows the mean relative error $\lvert (\widehat Z_j/\widehat Z_i)/(Z_j/Z_i)-1\rvert$ for a pair of frames $i<j$ across simulated clips. For a tracked sphere moving on a three-dimensional straight line under a fixed pinhole camera, the image displacement from its first position multiplied by $Z(t)/Z(0)$ is proportional to distance traveled along that line. The derivation requires a common starting point. An interval between arbitrary image positions needs its own projection geometry.

For this single-ball case, a temporal head operating on frozen DA3 features~\cite{da3} proposes a relative-depth curve. Either sufficient evidence from the curve or an independent contour-width test can enable correction for the clip. Apparent width helps decide when to trust the curve. When neither test triggers, the applied multiplier stays at one. Camera motion, curved paths and non-spherical objects require their own geometry. At L4, two independently normalized depth tracks also leave the scale between objects undetermined. Other tasks use a verified measurement plane, a common scale or a projective invariant such as the cross-ratio of four collinear points.

\subsection{Matching the evaluator to the claim}
We distinguish four judgments. Process realization ($E$) identifies what happened to each designated object in the event window. Integrity ($I$) concerns continuity of the identities and task-relevant properties of those objects. Qualitative agreement ($Q$) is the fraction of prespecified observable conditions satisfied, for example the ordering of two contact times. Quantitative agreement ($P$) measures the magnitude of a physical residual. A valid bounding box can support localization while leaving all four judgments open. L0 and I use native 1--5 ratings; for L1 and L2 we use condition-specific judgments, and for L3 and L4 we recover the relevant physical quantities. Section~\ref{sec:experiments} reports the independent validation data available for each type of evidence.

For a positive observed ratio $\widehat R_u$ and theoretical ratio $R_u^\star$, the quantitative agreement of unit $u$ is
\begin{equation}
P_u = \exp\!\left[-\left|\log\!\left(\frac{\widehat R_u}{R_u^\star}\right)\right|\right].
\label{eq:unit-quantitative-agreement}
\end{equation}
Reciprocal proportional errors receive equal weight. Equal heights and other zero-valued targets use a prespecified normalized residual. We also report the physical measurement itself. In the depth experiments, we measure pairwise physics ratio error as
\begin{equation}
\mathrm{PPRE} = \left|1-\frac{R^\star}{R_{\mathrm{obs}}}\right|.
\label{eq:pairwise-physics-ratio-error}
\end{equation}
Agreement with a three-dimensional reference PPRE measures the error of the evaluator even when the trajectory itself departs from the ideal law.

Joint Physical Fidelity (JPF) combines evidence for the same objects and times. For fully observed units, a joint score requires calibrated factors $\widetilde E_u,\widetilde I_u\in[0,1]$ derived from the native 1--5 ratings, qualitative indicators $Q_u\in\{0,1\}$, and weights summing to one:
\begin{equation}
\mathrm{JPF} = 100\sum_u w_u \widetilde E_u \widetilde I_u Q_u P_u.
\label{eq:joint-physical-fidelity}
\end{equation}
Missing evidence retains its weight. We bound the joint score by allowing each unknown component to range from zero to one, and report physical measurement coverage separately. A confirmed failure can identify a zero joint contribution even when a distance cannot be measured. We report E and I on their native 1--5 scale. They are not binary gates; a joint point estimate requires an independently validated mapping to the normalized factors.

\section{Validating the Evaluators}
\label{sec:experiments}
A human rating can tell whether a fall looks plausible, but checking a depth estimate requires the object's measured 3D position. We validate each evaluator against an independent reference matching its benchmark conditions in scene content, event interval, and camera geometry.

\subsection{Event and object-persistence judgments}
Gemini 3.8 Flash judges whether an event follows the applicable physical laws ($E$) and whether the objects persist throughout the clip ($I$). PhyGround~\cite{phyground} provides separate human ratings for these judgments. For $E$, the reference is the mean rating of the laws applicable to that clip. For $I$, it is the mean human rating of object persistence over the full sequence.

Two videos can depict the same event but show different apparatus or viewpoints. We therefore match videos by both the task description and the appearance of the opening frame, giving priority to coverage across tasks and configurations. Under the same assignment rules, this procedure produces closer visual matches than random selection at equal task and configuration coverage (Fig.\ref{fig:evaluator-validation}a; Appendix Figs.\ref{fig:ei-task-coverage},~\ref{fig:ei-matches}).

\begin{table*}[!t]
\centering
\caption{\textbf{Physical consistency} ($\uparrow$). L0 ranges from 1 to 5 and L1--L4 from 0 to 1. Bold marks row maxima, gray the difference from the maximum. Predict2.5 14B and Cosmos3 Super denote Cosmos-Predict2.5-14B and Cosmos3-Super-Image2Video.}
\label{tab:generator-main}
\begingroup
\small
\setlength{\tabcolsep}{1pt}
\setlength{\fboxsep}{0.75pt}
\setlength{\aboverulesep}{0.75pt}
\setlength{\belowrulesep}{0.95pt}
\renewcommand{\arraystretch}{0.81}
\newcommand{\familybest}[1]{\colorbox[rgb]{0.86,0.93,0.94}{\bfseries #1}}
\newcommand{\familyscore}[2]{#1\hspace{1.4pt}{\scriptsize\color[gray]{0.33}$-#2$}}
\newcommand{\familylabel}[2]{\multirow{5}{*}{\shortstack[l]{\makebox[0pt][l]{\includegraphics[page=#1,width=63pt]{table4_scene_symbols.pdf}}\\[0.5pt]#2}}}
\begin{tabular*}{\textwidth}{@{\extracolsep{\fill}}ll|*{8}{c}@{}}
\toprule[0.85pt]
\multicolumn{2}{c}{\bfseries Evaluation} & \multicolumn{8}{c}{\bfseries Video generators} \\
\cmidrule(lr){1-2}\cmidrule(l){3-10}
\textbf{Physics family} & \textbf{Level} & {\footnotesize\bfseries\shortstack{Pyramid Flow\\(SD3)\, {\normalfont\scriptsize\cite{jin2025pyramidflow}}}} & {\footnotesize\bfseries\shortstack{Cosmos-Predict1\\14B\, {\normalfont\scriptsize\cite{cosmosp1video}}}} & {\footnotesize\bfseries\shortstack{Cosmos-Predict2\\14B\, {\normalfont\scriptsize\cite{cosmosp2video}}}} & {\footnotesize\bfseries\shortstack{Predict2.5\\14B\, {\normalfont\scriptsize\cite{cosmosp25video}}}} & {\footnotesize\bfseries\shortstack{Wan2.2-I2V\\A14B\, {\normalfont\scriptsize\cite{wan22i2v}}}} & {\footnotesize\bfseries\shortstack{Cosmos3\\Super\, {\normalfont\scriptsize\cite{cosmos3supervideo}}}} & {\footnotesize\bfseries\shortstack{MiniMax-H3\\Highspeed\, {\normalfont\scriptsize\cite{minimaxh3highspeed}}}} & {\footnotesize\bfseries\shortstack{Seedance\\2.5\, {\normalfont\scriptsize\cite{bytedanceseed25}}}} \\
\midrule[0.5pt]
\familylabel{1}{Motion and\\energy} & L0 & \familyscore{2.48}{1.02} & \familyscore{2.00}{1.50} & \familyscore{2.14}{1.36} & \familyscore{2.43}{1.07} & \familyscore{2.72}{0.78} & \familyscore{2.58}{0.92} & \familyscore{3.44}{0.06} & \familybest{3.50} \\
 & L1 & \familyscore{0.00}{0.57} & \familyscore{0.00}{0.57} & \familyscore{0.11}{0.46} & \familyscore{0.20}{0.37} & \familyscore{0.29}{0.28} & \familyscore{0.48}{0.09} & \familybest{0.57} & \familyscore{0.56}{0.01} \\
 & L2 & \familyscore{0.04}{0.73} & \familyscore{0.00}{0.77} & \familyscore{0.00}{0.77} & \familyscore{0.21}{0.56} & \familyscore{0.23}{0.54} & \familyscore{0.43}{0.34} & \familyscore{0.67}{0.10} & \familybest{0.77} \\
 & L3 & \familyscore{0.35}{0.39} & \familyscore{0.41}{0.33} & \familyscore{0.45}{0.29} & \familyscore{0.50}{0.24} & \familyscore{0.52}{0.22} & \familyscore{0.56}{0.18} & \familyscore{0.71}{0.03} & \familybest{0.74} \\
 & L4 & \familyscore{0.28}{0.39} & \familyscore{0.32}{0.35} & \familyscore{0.45}{0.22} & \familyscore{0.48}{0.19} & \familyscore{0.51}{0.16} & \familyscore{0.57}{0.10} & \familyscore{0.62}{0.05} & \familybest{0.67} \\
\midrule[0.30pt]
\familylabel{2}{Oscillation\\and statics} & L0 & \familyscore{3.06}{0.75} & \familyscore{3.00}{0.81} & \familyscore{3.08}{0.73} & \familyscore{3.16}{0.65} & \familyscore{3.33}{0.48} & \familyscore{3.00}{0.81} & \familyscore{3.72}{0.09} & \familybest{3.81} \\
 & L1 & \familyscore{0.00}{0.50} & \familyscore{0.00}{0.50} & \familyscore{0.00}{0.50} & \familyscore{0.07}{0.43} & \familyscore{0.14}{0.36} & \familyscore{0.00}{0.50} & \familyscore{0.43}{0.07} & \familybest{0.50} \\
 & L2 & \familyscore{0.15}{0.41} & \familyscore{0.40}{0.16} & \familyscore{0.18}{0.38} & \familyscore{0.19}{0.37} & \familyscore{0.20}{0.36} & \familyscore{0.25}{0.31} & \familyscore{0.48}{0.08} & \familybest{0.56} \\
 & L3 & \familyscore{0.32}{0.63} & \familyscore{0.35}{0.60} & \familyscore{0.45}{0.50} & \familyscore{0.49}{0.46} & \familyscore{0.49}{0.46} & \familyscore{0.52}{0.43} & \familybest{0.95} & \familyscore{0.94}{0.01} \\
 & L4 & \familyscore{0.30}{0.42} & \familyscore{0.28}{0.44} & \familyscore{0.40}{0.32} & \familyscore{0.42}{0.30} & \familyscore{0.46}{0.26} & \familyscore{0.55}{0.17} & \familyscore{0.63}{0.09} & \familybest{0.72} \\
\midrule[0.30pt]
\familylabel{3}{Fluids} & L0 & \familyscore{2.48}{1.00} & \familyscore{1.96}{1.52} & \familyscore{1.98}{1.50} & \familyscore{2.00}{1.48} & \familyscore{2.00}{1.48} & \familyscore{2.58}{0.90} & \familyscore{3.31}{0.17} & \familybest{3.48} \\
 & L1 & \familyscore{0.00}{0.71} & \familyscore{0.51}{0.20} & \familyscore{0.45}{0.26} & \familyscore{0.43}{0.28} & \familyscore{0.40}{0.31} & \familyscore{0.25}{0.46} & \familyscore{0.62}{0.09} & \familybest{0.71} \\
 & L2 & \familyscore{0.14}{0.39} & \familyscore{0.00}{0.53} & \familyscore{0.14}{0.39} & \familyscore{0.11}{0.42} & \familyscore{0.07}{0.46} & \familyscore{0.36}{0.17} & \familyscore{0.37}{0.16} & \familybest{0.53} \\
 & L3 & \familyscore{0.30}{0.46} & \familyscore{0.38}{0.38} & \familyscore{0.38}{0.38} & \familyscore{0.51}{0.25} & \familyscore{0.57}{0.19} & \familyscore{0.62}{0.14} & \familyscore{0.65}{0.11} & \familybest{0.76} \\
 & L4 & \familyscore{0.26}{0.37} & \familyscore{0.31}{0.32} & \familyscore{0.34}{0.29} & \familyscore{0.36}{0.27} & \familyscore{0.42}{0.21} & \familyscore{0.43}{0.20} & \familyscore{0.59}{0.04} & \familybest{0.63} \\
\midrule[0.30pt]
\familylabel{4}{Optics and\\geometry} & L0 & \familyscore{3.00}{0.20} & \familyscore{1.87}{1.33} & \familyscore{1.89}{1.31} & \familyscore{1.92}{1.28} & \familyscore{1.91}{1.29} & \familyscore{2.49}{0.71} & \familyscore{2.93}{0.27} & \familybest{3.20} \\
 & L1 & \familyscore{0.00}{0.50} & \familyscore{0.00}{0.50} & \familyscore{0.00}{0.50} & \familyscore{0.01}{0.49} & \familyscore{0.01}{0.49} & \familyscore{0.00}{0.50} & \familyscore{0.40}{0.10} & \familybest{0.50} \\
 & L2 & \familyscore{0.33}{0.20} & \familyscore{0.36}{0.17} & \familyscore{0.19}{0.34} & \familyscore{0.20}{0.33} & \familyscore{0.22}{0.31} & \familyscore{0.42}{0.11} & \familybest{0.53} & \familyscore{0.42}{0.11} \\
 & L3 & \familyscore{0.33}{0.27} & \familyscore{0.42}{0.18} & \familyscore{0.44}{0.16} & \familyscore{0.52}{0.08} & \familyscore{0.50}{0.10} & \familyscore{0.55}{0.05} & \familyscore{0.58}{0.02} & \familybest{0.60} \\
 & L4 & \familyscore{0.30}{0.27} & \familyscore{0.35}{0.22} & \familyscore{0.31}{0.26} & \familyscore{0.31}{0.26} & \familyscore{0.37}{0.20} & \familyscore{0.40}{0.17} & \familyscore{0.54}{0.03} & \familybest{0.57} \\
\midrule[0.30pt]
\familylabel{5}{Electromagnetism} & L0 & \familyscore{3.08}{0.68} & \familyscore{3.13}{0.63} & \familyscore{3.16}{0.60} & \familyscore{3.18}{0.58} & \familyscore{3.35}{0.41} & \familyscore{3.49}{0.27} & \familyscore{3.74}{0.02} & \familybest{3.76} \\
 & L1 & \familyscore{0.00}{1.00} & \familyscore{0.69}{0.31} & \familyscore{0.69}{0.31} & \familyscore{0.73}{0.27} & \familyscore{0.82}{0.18} & \familyscore{0.91}{0.09} & \familybest{1.00} & \familybest{1.00} \\
 & L2 & \familyscore{0.34}{0.66} & \familyscore{0.53}{0.47} & \familyscore{0.54}{0.46} & \familyscore{0.61}{0.39} & \familyscore{0.69}{0.31} & \familyscore{0.76}{0.24} & \familyscore{0.84}{0.16} & \familybest{1.00} \\
 & L3 & \familyscore{0.36}{0.31} & \familyscore{0.40}{0.27} & \familyscore{0.38}{0.29} & \familyscore{0.40}{0.27} & \familyscore{0.39}{0.28} & \familyscore{0.41}{0.26} & \familyscore{0.62}{0.05} & \familybest{0.67} \\
 & L4 & \familyscore{0.32}{0.28} & \familyscore{0.36}{0.24} & \familyscore{0.35}{0.25} & \familyscore{0.34}{0.26} & \familyscore{0.39}{0.21} & \familyscore{0.39}{0.21} & \familybest{0.60} & \familybest{0.60} \\
\midrule[0.30pt]
\familylabel{6}{Thermal physics} & L0 & \familyscore{2.44}{0.61} & \familyscore{1.91}{1.14} & \familyscore{1.94}{1.11} & \familyscore{1.96}{1.09} & \familyscore{1.96}{1.09} & \familyscore{2.54}{0.51} & \familyscore{3.00}{0.05} & \familybest{3.05} \\
 & L1 & \familyscore{0.00}{1.00} & \familyscore{0.69}{0.31} & \familyscore{0.69}{0.31} & \familyscore{0.73}{0.27} & \familyscore{0.82}{0.18} & \familyscore{0.91}{0.09} & \familybest{1.00} & \familybest{1.00} \\
 & L2 & \familyscore{0.35}{0.60} & \familyscore{0.55}{0.40} & \familyscore{0.56}{0.39} & \familyscore{0.64}{0.31} & \familyscore{0.71}{0.24} & \familyscore{0.79}{0.16} & \familyscore{0.87}{0.08} & \familybest{0.95} \\
 & L3 & \familyscore{0.60}{0.34} & \familyscore{0.65}{0.29} & \familyscore{0.68}{0.26} & \familyscore{0.68}{0.26} & \familyscore{0.79}{0.15} & \familyscore{0.82}{0.12} & \familyscore{0.93}{0.01} & \familybest{0.94} \\
 & L4 & \familyscore{0.57}{0.38} & \familyscore{0.62}{0.33} & \familyscore{0.64}{0.31} & \familyscore{0.65}{0.30} & \familyscore{0.77}{0.18} & \familyscore{0.81}{0.14} & \familyscore{0.93}{0.02} & \familybest{0.95} \\
\bottomrule[0.85pt]
\end{tabular*}
\endgroup
\end{table*}

\subsection{Qualitative relations}
Each judge sees the same ordered frames and decides which specified qualitative conditions hold. The video score $Q$ is the fraction judged to hold. We match each condition to a PhysBench annotation~\cite{physbench} or a PhysInOne simulator state~\cite{physinone} for the interval in which the event is visible. The judges do not see these references. Direction, trend, ordering, and stability have the clearest coverage on the external data (Appendix Fig.~\ref{fig:q-coverage}).

\begin{figure}[!htbp]
\centering
\includegraphics[width=\columnwidth]{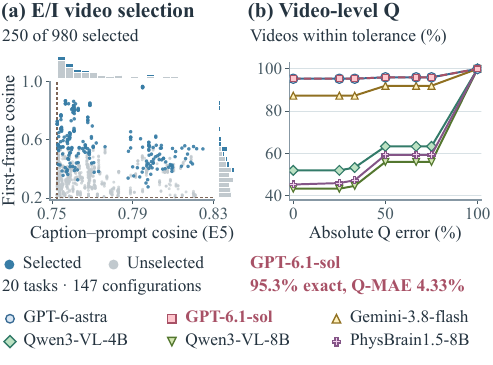}
\caption{(a) Text and first-frame similarity to benchmark scenes for selected and remaining PhyGround videos. (b) Fraction of videos within each tolerance for absolute $Q$ error.}
\label{fig:evaluator-validation}
\end{figure}

We compare each predicted $Q$ with its reference and give each video equal weight. GPT-6.1-sol matches the reference score exactly on 95.3\% of videos and ties GPT-6-astra for the lowest mean absolute error. Its lower inference cost makes it our choice for scoring qualitative relations. Figure~\ref{fig:evaluator-validation}b compares the judges across error tolerances, and Fig.~\ref{fig:condition-span}a breaks down errors by relation.

\subsection{Quantitative measurement under depth change}
\setcounter{dbltopnumber}{2}
We test our curve--contour evaluator on straight ball tracks from fixed cameras. Simulated trajectories and ABCD motion capture~\cite{abcd} provide independent positions from which we compute cross-frame depth ratios. On both sources, our estimates have lower error than the fixed-depth baseline. The simulated trajectories also let us compute the displacement-ratio error directly in 3D. Correcting the image measurements brings their error closer to this reference, which is the quantity that matters for L3 and L4.

\begin{figure}[!hb]
\centering
\includegraphics[width=\columnwidth]{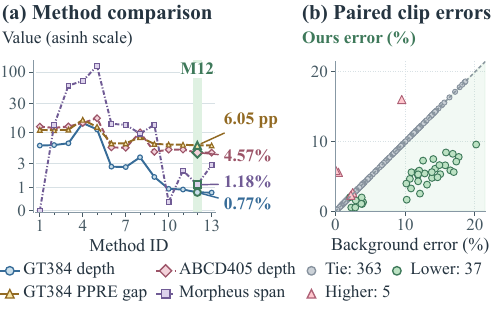}
\caption{(a) Depth and displacement-ratio metrics by method. (b) Paired motion-capture errors for our curve--contour evaluator (M12) and the background-based rule. The diagonal marks equal error.}
\label{fig:depth-methods}
\end{figure}

\begin{figure}[tbp]
\centering
\includegraphics[width=\columnwidth]{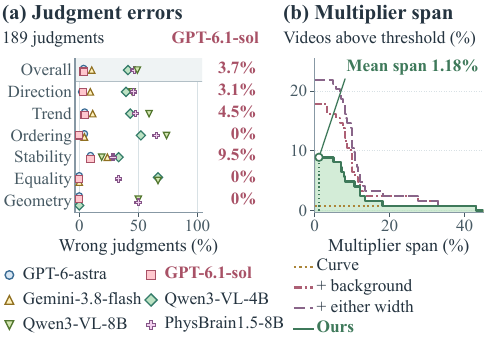}
\caption{(a) Qualitative judgment errors by relation. (b) Within-video variation of the depth correction on Morpheus. The marker gives the mean span for our evaluator (M12).}
\label{fig:condition-span}
\end{figure}

To isolate the decision to apply a correction, we keep the depth predictor fixed and compare the two decision rules on the same motion-captured tracks. On Morpheus footage~\cite{morpheus}, we  measure how much the correction factor varies within a video (Fig.~\ref{fig:condition-span}b). Appendix Table~\ref{tab:depth-method-intervals} and Fig.~\ref{fig:depth-gate-paired} give the method comparisons. The separate audit of refraction angles appears in Appendix~\ref{sec:historical}.

\section{Experiments}
\label{sec:generator-experiments}

Each model receives a first frame and a prompt. \Cref{tab:generator-main} compares eight generators across six physics families. L0 uses applicable law ratings as a proxy for event consistency, with persistence scored separately. L1 tests paired interventions, and L2 compares instances within a clip. L3 and L4 test measured ratios against physical predictions (\cref{fig:fall-levels}). Coverage and run settings are given in Appendix~\ref{app:generator-settings}.

\subsection{Main Results}
\label{sec:main-results}
\noindent\textbf{(1) The two proprietary models lead on qualitative consistency and already reproduce some quantitative relations.} Averaged equally over models and physics families, the two proprietary models score 3.41 out of 5 at L0, versus 2.53 for the six open-weight models. The gaps persist at L1 (0.69 vs. 0.31) and L2 (0.67 vs. 0.33). The two proprietary models also average 0.76 at L3 and 0.67 at L4.

\smallskip\noindent\textbf{(2) Between-instance comparisons are harder.} L4 scores fall below L3 in 42 of 48 model--family comparisons, tie in three and exceed L3 in three. All eight models score lower at L4 on average. In optics and geometry the gap narrows from 0.15 for Cosmos3-Super-Image2Video to 0.04 for MiniMax and 0.03 for Seedance. In oscillation and statics, the two proprietary models still have gaps of 0.32 and 0.22.

\smallskip\noindent\textbf{(3) Performance varies by physical scene.} Electromagnetism and thermal physics both average 0.73 at L1. Thermal physics also leads at L3 (0.76) and L4 (0.74). A dial, needle or piston shows the response in many of these tasks and may be easier to render than a changing light path. Oscillation and statics scores well at L0 (3.27 out of 5). Familiar motions such as pendulum swings may be more common in training footage, although we did not measure their frequency. \textbf{Optics and geometry scores lowest at L1 (0.12). Models struggle to reproduce how changing the incident ray affects reflection and refraction.}

\par\smallskip\noindent
\textbf{(4) Depth changes affect quantitative scores.} The larger a track's depth excursion, the more its L3 and L4 scores change after perspective correction (\cref{fig:generator-depth-effect}a). This trend shows that motion in depth does bias the uncorrected scores, and that the correction acts where this bias is largest. Combined with its accuracy on ABCD motion capture (12.5\% → 4.6\% depth-ratio error), this indicates the correction is effective on generated videos.

\smallskip\noindent\textbf{(5) Agreement often peaks in the middle window.} For both MiniMax-H3 and Seedance 2.5, mean agreement rises from the early to the middle window for each of the four laws. Six of the eight model--law curves then decline in the late window, including all four curves for uniform motion and constant acceleration (\cref{fig:generator-depth-effect}b).

\par\smallskip\noindent\begin{minipage}{\columnwidth}
\centering
\includegraphics[width=\linewidth]{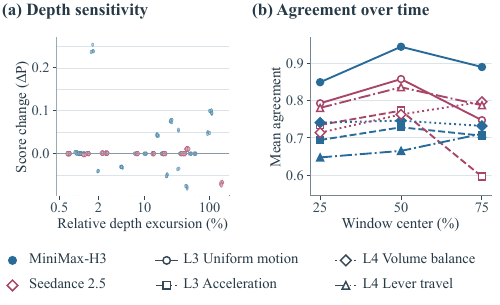}
\captionsetup{hypcap=false}
\captionof{figure}{(a) Change in L3 and L4 agreement after depth correction, $\Delta P=P_{\mathrm{corrected}}-P_{\mathrm{raw}}$. Positive values indicate closer agreement with the law. (b) Mean agreement with four laws in windows centered at 25\%, 50\% and 75\% of the measured phase.}
\label{fig:generator-depth-effect}
\end{minipage}\par

\smallskip\noindent\textbf{(6) Both models underestimate differences between objects.} Objects in the same scene showed less contrast in their physical responses than predicted ($G<1$) in 77.2\% of MiniMax-H3 videos and 73.7\% of Seedance 2.5 videos (\cref{fig:response-contrast}). MiniMax-H3 compressed the differences more strongly, with median contrast retention of 0.168 versus 0.330 for Seedance 2.5.

\par\smallskip\noindent\begin{minipage}{\columnwidth}
\centering
\includegraphics[width=.492\linewidth]{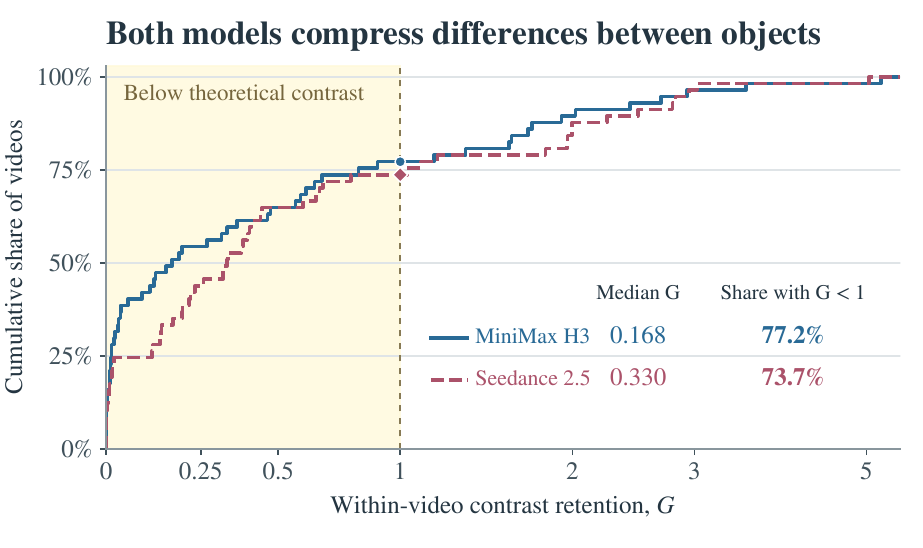}\hfill
\includegraphics[width=.492\linewidth]{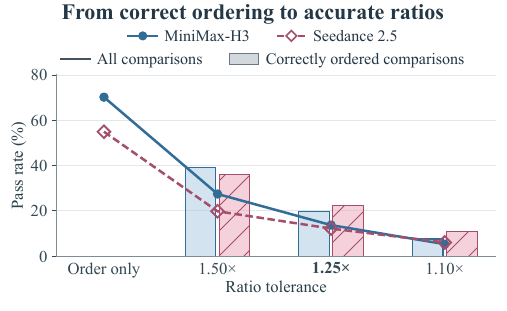}
\captionsetup{hypcap=false}
\captionof{figure}{Left: distribution of per-video median gain $G=|\log R_{\mathrm{obs}}|/|\log R_{\mathrm{theory}}|$ for object response ratios $R$. Right: pass rates at each tolerance for all pairs (lines) and correctly ordered pairs (bars).}
\label{fig:response-contrast}
\end{minipage}\par

\section{Discussion}
A plausible event can still have the wrong physical ratios. Reporting measurement coverage alongside each score distinguishes these errors from comparisons the video cannot support. Independent checks are strongest for common qualitative relations, such as direction and ordering, and for relative depth along straight ball trajectories.

\section*{Limitations}
Our five levels describe a progression from qualitative to quantitative physical consistency, but some scenes support only a subset of them. A single pendulum allows us to test whether its period stays constant across cycles (L3). Testing how the period ratio depends on relative length (L4) requires a second pendulum in the same scene. Even when adjacent levels both apply, the gap in difficulty can vary by phenomenon. Finer criteria within each level could capture these scene-specific differences. Our RGB tasks span six physics families and leave astrophysical phenomena such as planetary orbits untested.

\section*{LLM Usage Statement}
We used large language model (LLM) tools during the preparation of this work for three purposes: (1) assisting with code refinement and runtime efficiency optimization; (2) conducting preliminary background research and organizing relevant prior work; and (3) polishing the language of the manuscript after the content had been written by the authors. All LLM-generated suggestions were critically reviewed, verified, and revised by the authors. The authors take full responsibility for the methodology, experiments, results, claims, citations, and final content of this paper.

\clearpage

{\small
\bibliographystyle{ieeenat_fullname}
\bibliography{physq_references_comparison,paper_eval_sources,related_work_sources}
}
\clearpage
\appendix
\setcounter{dbltopnumber}{3}
\twocolumn[{
\section{Input Construction and Evaluation Procedure}
\label{app:construction-evaluation}
\begin{minipage}{\textwidth}
\captionsetup{type=figure}
\centering
\includegraphics[width=\textwidth]{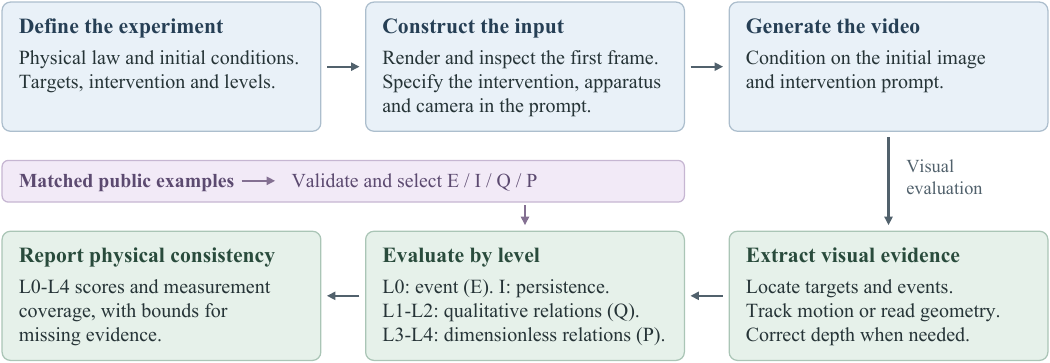}
\caption{Data construction and visual evaluation. Initial images and intervention prompts define the generation inputs. Evaluators validated on matched public examples support qualitative judgments and law-derived dimensionless measurements, with depth correction when needed. Scores are reported with measurement coverage and bounds for missing evidence.}
\label{fig:construction-evaluation}
\end{minipage}
\par\medskip
\begin{minipage}{\textwidth}
\captionsetup{type=table}
\caption{Evaluators used at each physical-consistency level, their validation datasets, and the corresponding evidence.}
\label{tab:evaluator-level-map}
\centering\small
\setlength{\tabcolsep}{3pt}
\renewcommand{\arraystretch}{1.12}
\begin{tabularx}{\textwidth}{@{}p{0.04\textwidth}>{\raggedright\arraybackslash}p{0.20\textwidth}>{\raggedright\arraybackslash}p{0.23\textwidth}>{\raggedright\arraybackslash}p{0.23\textwidth}>{\raggedright\arraybackslash}X@{}}
\toprule
Type & Role in L0--L4 & Selected evaluator & Validation data & Validation evidence \\
\midrule
$E$ & L0: event consistency & Gemini 3.8 Flash & PhyGround~\cite{phyground}: physical-law annotations & Matching: Fig.~\ref{fig:evaluator-validation}(a). Coverage: Figs.~\ref{fig:ei-task-coverage}(c),~\ref{fig:ei-matches}. \\
\addlinespace[3pt]
$I$ & L0--L4: object integrity and persistence, assessed separately & Gemini 3.8 Flash & PhyGround~\cite{phyground}: object-persistence annotations & Matching: Fig.~\ref{fig:evaluator-validation}(a). Coverage: Figs.~\ref{fig:ei-task-coverage}(c),~\ref{fig:ei-matches}. \\
\addlinespace[3pt]
$Q$ & L1: response to intervention; L2: relations between instances & GPT-6.1-sol & PhysBench~\cite{physbench} annotations; PhysInOne~\cite{physinone} simulator states & Errors: Figs.~\ref{fig:evaluator-validation}(b),~\ref{fig:condition-span}(a). Coverage: Fig.~\ref{fig:q-coverage}. \\
\addlinespace[3pt]
$P$ & L3: quantitative agreement within an instance; L4: quantitative agreement between instances & GPT-6.1-sol localization; NanoTrack v2 tracking; task-specific measurements; Curve--Contour Depth Evaluator (M12) for depth correction & ABCD~\cite{abcd}: motion-capture depth; GT384: simulated 3D reference; Morpheus~\cite{morpheus}: correction stability, without depth ground truth & Depth: Figs.~\ref{fig:depth-methods},~\ref{fig:depth-gate-paired}(a,b). Stability: Fig.~\ref{fig:condition-span}(b). Metrics: Table~\ref{tab:depth-method-intervals}. \\
\bottomrule
\end{tabularx}
\end{minipage}
\par\medskip
}]

\subsection{Data construction}
\label{app:input-record}
The benchmark contains 594 inputs spanning 23 tasks in six physics families and supports 1,231 evaluation samples.

The evaluation samples comprise 355 at L0, 99 at L1, 252 at L2, 305 at L3 and 220 at L4, matching \cref{fig:inventory}(a). Each L1 comparison is a distinct controlled pair; the 99 pairs involve 183 inputs. At the other levels, an input contributes one comparison per applicable level.

Each configuration specifies the apparatus, initial state, intervention and expected physical relation. We pair it with one or more background variants and define the applicable levels before generation. Each generation input consists of an initial image and a generation prompt, together with the event, object-persistence and qualitative criteria used for evaluation. Physical laws determine the quantitative targets.

\begin{table}[!ht]
\centering\small
\caption{Five levels of physical consistency. Each criterion applies across physical settings. \Cref{fig:fall-levels} illustrates all five with free fall.}
\label{tab:levels}
\begin{tabularx}{\columnwidth}{@{}lX@{}}
\toprule
Level & Physical consistency criterion \\
\midrule
L0 & \textbf{Event consistency.} The physical event expected under the stated conditions occurs. \\
L1 & \textbf{Qualitative response consistency.} Across matched trials, the same instance responds to a controlled change as physics predicts. \\
L2 & \textbf{Qualitative relational consistency.} Distinct instances in one scene exhibit the predicted qualitative relation. \\
L3 & \textbf{Within-instance quantitative consistency.} Quantities measured within one instance satisfy the numerical relation predicted by the law. \\
L4 & \textbf{Between-instance quantitative consistency.} Measurements from distinct instances in one scene satisfy the numerical relation predicted by the law. \\
\bottomrule
\end{tabularx}
\end{table}

We use the source image as the initial frame or revise it through controlled rerendering and visual inspection. The generation prompt specifies the initial conditions, intervention, camera constraints and observation window. For visual evaluation, the default generation setting is a five-second, 480p clip with the input aspect ratio and no audio; evaluation preserves the model's native frame rate.

\subsection{Evaluation procedure}
\label{app:evaluation-record}
We evaluate the generated video directly, retaining its original frame timing without interpolation or additional lossy compression. E checks whether the expected event occurs, I checks object persistence, and Q checks qualitative relations. L1 compares matched trials under different interventions; L2 compares instances within one video. L3 and L4 test quantitative relations within and between instances, respectively.

Quantitative evaluation first identifies the target objects, relevant events and measurable geometry. Tracking and task-specific image measurements then recover the quantities needed to test dimensionless physical relations. When depth changes affect measurement, we detect and correct them before scoring.

Before benchmark evaluation, we test candidate evaluators on public examples selected for similarity in task semantics, objects and viewing conditions. Comparison with independent annotations or known geometry supports evaluator selection for the corresponding level.

\subsection{A generation prompt and its evaluation questions}
\label{app:prompt-example}
Generation prompt for the free-fall example:

\begin{quote}\small
Use the input image as the first frame and retain its background, materials and composition. Preserve the initial positions, relative heights, sizes and colors of the red, blue and green balls, and the plate or horizontal contact surface below them. Keep the release heights shown in the image. All three balls start at rest.

The first frame shows their common release time. Release all three balls simultaneously from rest in a single trial, without an extra push. Record each fall and first contact continuously from release, using the initial heights and relative positions in the image.

Until a ball first touches the plate below it, it experiences only downward gravity in the same uniform field. Neglect air resistance and impart no release impulse. Keep the plate or contact surface fixed.

Keep the three centers in the vertical plane shown in the first frame at a common, constant optical depth. Preserve each ball's spherical shape, physical size, color and identity, without stretching, damage, splitting, merging or replacement. Do not depict displacement by changing a ball's size.

Lock the camera to the first-frame view. Keep its position, orientation, focal length, exposure, white balance and framing fixed, without tracking, rotation or zoom. Record one uninterrupted five-second shot at constant frame rate and normal speed, without cuts, speed changes, slow motion or freeze frames. Use the 480p setting and retain the input aspect ratio and apparatus. Do not delay, slow down or repeat the experiment to fill the available time.

Keep the release positions, fall paths and first contact locations in view. Record free fall through first contact without hovering, resetting or releasing a ball again before contact. Do not stage an additional rebound demonstration. Preserve the apparatus, materials, background and physical markers under stable lighting. Add no people, hands, unrelated apparatus, obstructions, subtitles, formulas, measurements, arrows, rulers or drawn trajectories.
\end{quote}

At L0 the rubric asks whether each designated ball visibly falls through the air. It uses the frames before the first target makes contact and records an unreadable target separately from one that visibly stays in place. At L2 the registered question compares the first contacts of the individual balls with their order of initial clearance above the same plate, given simultaneous release from rest. At L3 the evaluator follows each ball from release within the common pre-contact interval. At L4, each ball is measured at its last descending state before rebound, whether or not contact is visible. The elapsed-time ratio is compared with $t_i/t_j=\sqrt{h_i/h_j}$, where $h_i$ and $h_j$ are the distances fallen from release to those states. Rebounds fall outside all four questions.

The Q evaluation checks six conditions: the direction and speed trend of each ball. For example, one condition asks whether the red ball's speed increases as it falls. Each judgment is supported by a video frame and a brief explanation. The score is the fraction of conditions satisfied. The L2 contact-order question uses the same judgment format over its relevant observation window.

The native PhyGround validation prompt asks for a 1--5 rating for each applicable physical law or for object persistence. In the free-fall example the native E question concerns gravity, while the native I question concerns persistence throughout the clip. These ratings retain the original anchors for comparison with public human annotations. L0 uses the mean of the applicable E ratings on the native 1--5 scale; I uses the same 1--5 scale for object persistence.

\begin{figure}[t]
\centering
\includegraphics[width=\columnwidth]{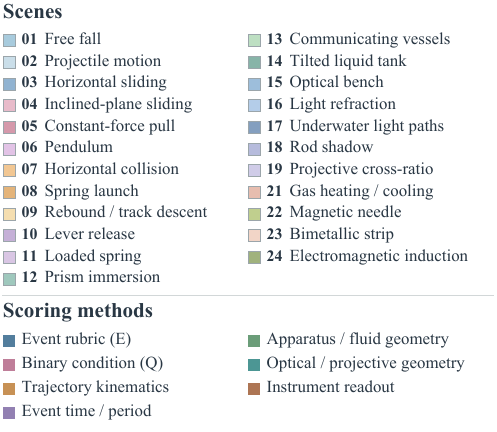}
\caption{Full key to Fig.~\ref{fig:inventory}. Numbers identify scenes in the middle ring. The seven scoring methods are color-coded in the outer ring.}
\label{fig:inventory-key}
\end{figure}

\subsection{Recovering quantities from the video}
\begin{table*}[t]
\centering
\caption{Depth evaluator means and 95\% bootstrap confidence intervals (in brackets). Depth error is the cross-frame depth-ratio error. The PPRE gap is the absolute difference between corrected and ground-truth physical ratio errors, in percentage points (pp). On Morpheus123, which lacks depth ground truth, span measures variation in the applied multiplier. The bootstrap uses 5,000 resamples. We resample GT384 clips within strata defined by trajectory and depth constancy, ABCD405 by source recording (115 recordings), and Morpheus123 by video after averaging targets within each video. Bold identifies the selected evaluator.}
\label{tab:depth-method-intervals}
\begingroup
\fontsize{8}{9.4}\selectfont
\setlength{\tabcolsep}{1.2pt}
\renewcommand{\arraystretch}{1.05}
\begin{tabular*}{\textwidth}{@{\extracolsep{\fill}}llrrrr@{}}
\toprule
 & & \multicolumn{2}{c}{GT384} & \multicolumn{1}{c}{ABCD405} & \multicolumn{1}{c}{Morpheus123} \\
\cmidrule(lr){3-4}\cmidrule(lr){5-5}\cmidrule(l){6-6}
ID & Method & Depth error (\%) & PPRE gap (pp) & Depth error (\%) & Multiplier span (\%) \\
\midrule
M1 & Fixed depth & 6.010 [5.124, 6.918] & 10.999 [9.084, 13.121] & 12.465 [11.319, 13.779] & 0.000 [0.000, 0.000] \\
M2 & DA3-Base & 6.131 [5.381, 6.913] & 10.849 [9.113, 12.818] & 11.868 [10.534, 13.448] & 13.170 [11.132, 15.485] \\
M3 & VDA-Metric Base & 6.572 [5.821, 7.352] & 11.060 [9.348, 13.011] & 12.545 [11.175, 14.065] & 59.702 [45.666, 76.474] \\
M4 & Depth Pro & 14.261 [13.140, 15.427] & 16.160 [13.930, 18.655] & 14.328 [12.941, 15.885] & 71.417 [55.434, 90.875] \\
M5 & UniDepthV2 ViT-L & 11.218 [10.425, 12.096] & 12.330 [10.458, 14.513] & 17.184 [15.273, 19.257] & 125.868 [67.653, 200.002] \\
M6 & Inverse ball size & 2.616 [2.164, 3.095] & 6.551 [5.193, 8.203] & 5.611 [5.144, 6.110] & 13.869 [11.967, 16.003] \\
\addlinespace[1.2pt]
M7 & Per-frame residual MLP & 2.572 [2.149, 3.022] & 6.552 [5.206, 8.193] & 5.521 [5.060, 6.016] & 13.312 [11.535, 15.249] \\
M8 & Area-regularized temporal CNN & 3.811 [3.308, 4.323] & 8.313 [6.786, 10.107] & 10.000 [9.113, 11.027] & 9.538 [8.161, 11.007] \\
M9 & Ungated V2 temporal residual & 1.691 [1.435, 1.965] & 6.400 [5.061, 8.034] & 4.713 [4.289, 5.173] & 13.388 [11.664, 15.169] \\
\addlinespace[1.2pt]
M10 & Depth curve alone & 0.945 [0.803, 1.111] & 6.267 [4.898, 7.933] & 5.141 [4.658, 5.664] & 0.350 [0.000, 1.050] \\
M11 & Curve or background width & 0.911 [0.760, 1.093] & 6.205 [4.816, 7.879] & 5.141 [4.658, 5.664] & 2.218 [1.227, 3.392] \\
\textbf{M12} & \textbf{Curve--Contour Depth Evaluator (Ours)} & \textbf{0.769 [0.635, 0.928]} & \textbf{6.055 [4.711, 7.686]} & \textbf{4.567 [4.155, 5.020]} & \textbf{1.176 [0.472, 2.129]} \\
M13 & Curve or either width & 0.758 [0.621, 0.915] & 6.039 [4.697, 7.669] & 4.567 [4.155, 5.020] & 2.781 [1.731, 3.990] \\
\bottomrule
\end{tabular*}
\endgroup
\end{table*}

\begin{figure*}[t]
\centering
\includegraphics[width=\textwidth]{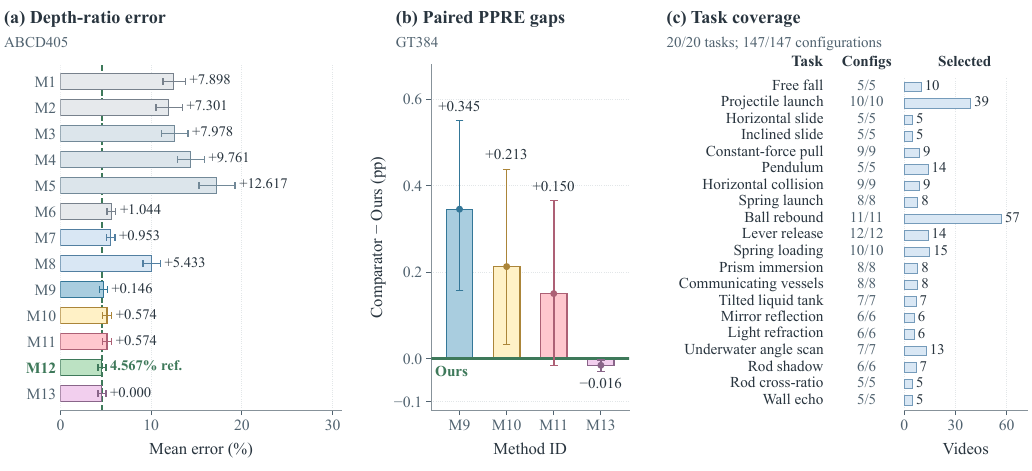}
\caption{Depth-evaluator comparison and PhyGround coverage. (a) Mean depth-ratio error on ABCD405; annotations give differences from Ours (M12) in percentage points. (b) Paired PPRE gaps on GT384 (comparator minus Ours); positive values favor Ours. Whiskers show 95\% bootstrap intervals; method IDs follow Table~\ref{tab:depth-method-intervals}. (c) Task coverage of 250 selected PhyGround videos: bars count videos, and fractions show represented configurations out of those considered.}
\label{fig:depth-gate-paired}
\label{fig:ei-task-coverage}
\end{figure*}

\label{app:recovering-quantities}
For tracked objects and markers, GPT-6.1 Sol locates each requested target in the first decoded frame. Its bounding box initializes a separate NanoTrack~v2 tracker, which returns a center per source frame when tracking succeeds. We retain the boxes and center coordinates to check for tracking drift. Liquid interfaces, ray directions and instrument readings use task-specific image measurements. For qualitative judgments, the level evaluator samples the registered event window at approximately 4\,Hz and retains its boundaries and frames adjacent to marked events. Paired trials are compared using the original video times.

The geometry and cutoff depend on the physical quantity. On an incline, each slider has a release frame, the end of its straight segment and its first arrival at the horizontal section. The straight-segment acceleration test ends before the bend. In communicating vessels, the evaluator measures liquid surfaces from a common vertical datum over a stable interval. For a ball in straight-line free fall under a fixed camera, the depth module accepts time-aligned frames and sphere masks and estimates depth relative to the first observation. The depth estimator uses a frozen DA3-Base model with a temporal head. It accepts timestamps, RGB frames and masks for one ball and returns both a candidate relative-depth curve and the multiplier actually applied. A frozen test of the curve or of single-frame contour width can enable the correction. Otherwise, the multiplier is one throughout the clip. Comparing different objects still requires independent evidence that their geometric scales agree. A missing landmark, unstable surface or unsupported scale is recorded with the affected measurement, alongside the number of comparisons that could be measured.

For a worked L4 example, three sliders on inclined tracks entered motion at different frames. All three remained visible on their straight tracks during frames 7--34 (1.125\,s). The red and blue sliders traveled 93.54 and 76.75 pixels during that common interval. After checking camera stability, apparent slider size and projected track directions, the measured red-to-blue displacement ratio was 1.219. The predicted ratio was 0.733 using the track angles, friction and each slider's elapsed time since release. The red and blue sliders separately took 65 and 49 frames from release to the bottom. Their time ratio, $65/49=1.33$, is observable directly. Testing the equal-drop prediction for those arrivals requires additional evidence that the two vertical drops agree. The video thus supports a time-ratio measurement even when its release geometry remains uncertain.

\subsection{Depth methods and correction gates}
\label{app:depth-methods}
Table~\ref{tab:depth-method-intervals} identifies the 13 methods in Fig.~\ref{fig:depth-methods}(a) and lists the four measurements in their original units. GT384 and ABCD405 supply reference depth. Morpheus123 supplies a measure of how much the applied correction varies within a video. A fixed-depth assumption gives zero span by construction. Figure~\ref{fig:depth-gate-paired}(a,b) compares the errors directly with the curve--contour gate.

\subsection{Coverage of the external evaluator tests}
\label{app:external-coverage}
\begin{figure*}[!tp]
\centering
\includegraphics[width=0.81\textwidth,trim=8bp 10bp 7bp 8bp,clip]{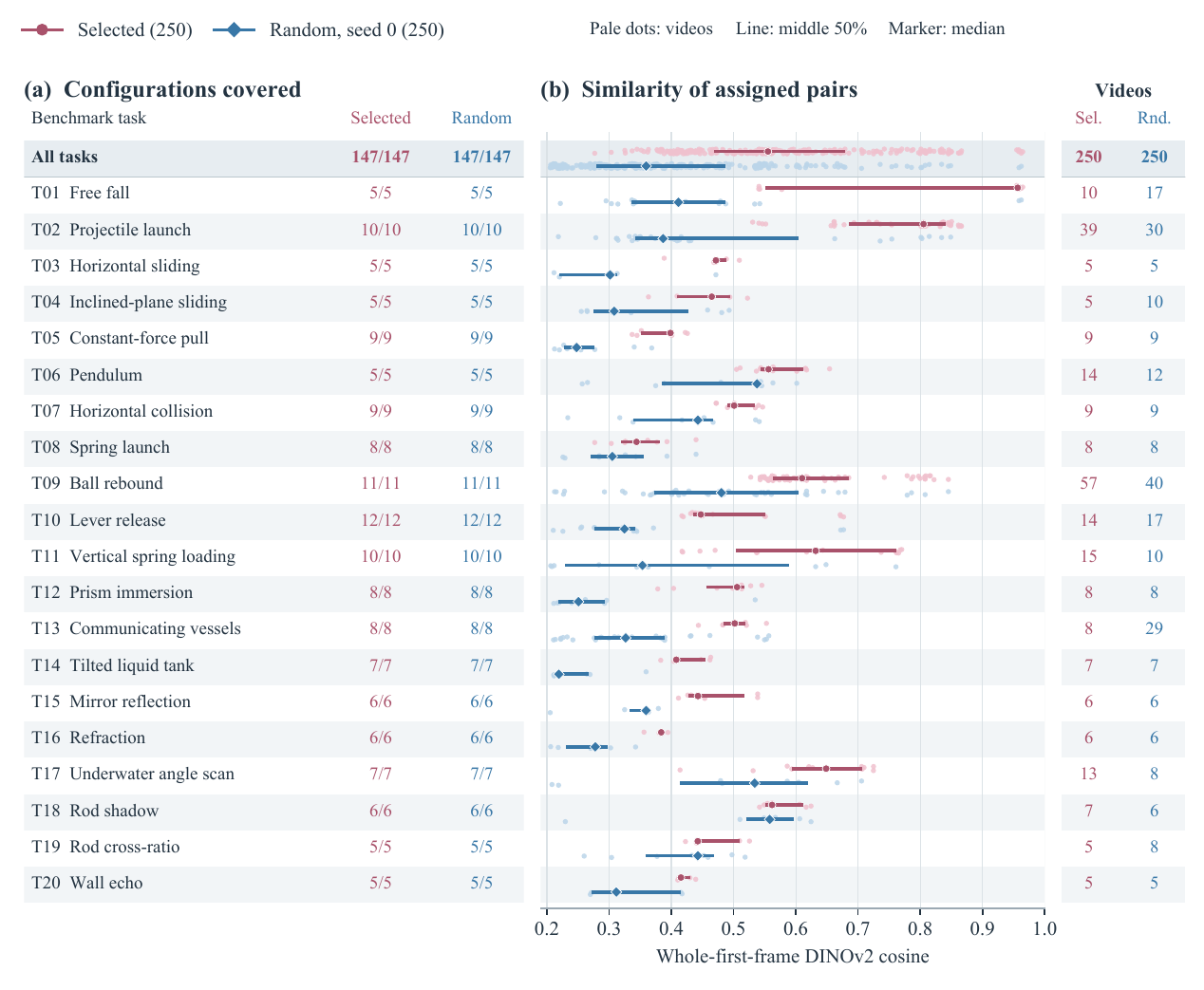}
\caption{Task-level comparison of selected and randomly drawn PhyGround videos. The two samples of 250 share 70 videos. Points show first-frame similarity under the same assignment rule. Both groups cover the 147 configurations used for this evaluator test; the selected videos are visually closer on average.}
\label{fig:ei-matches}
\end{figure*}

\begin{figure*}[!tp]
\centering
\includegraphics[width=\textwidth,trim=20bp 8bp 18bp 8bp,clip]{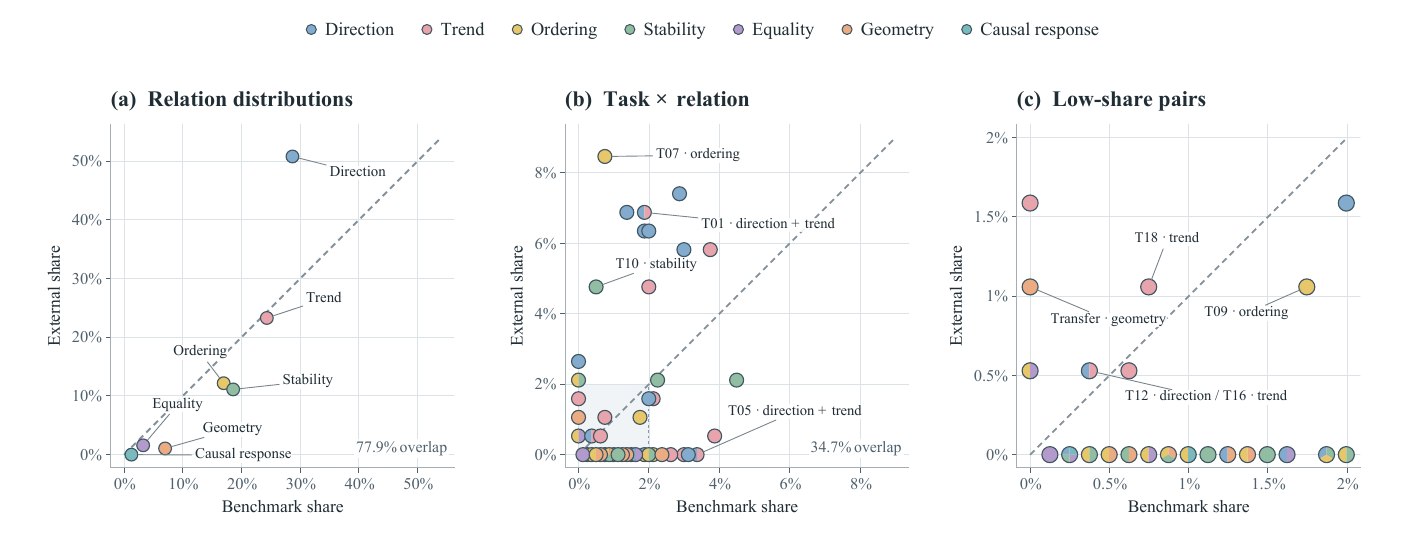}\par
\includegraphics[width=\textwidth,trim=24bp 8bp 12bp 4bp,clip]{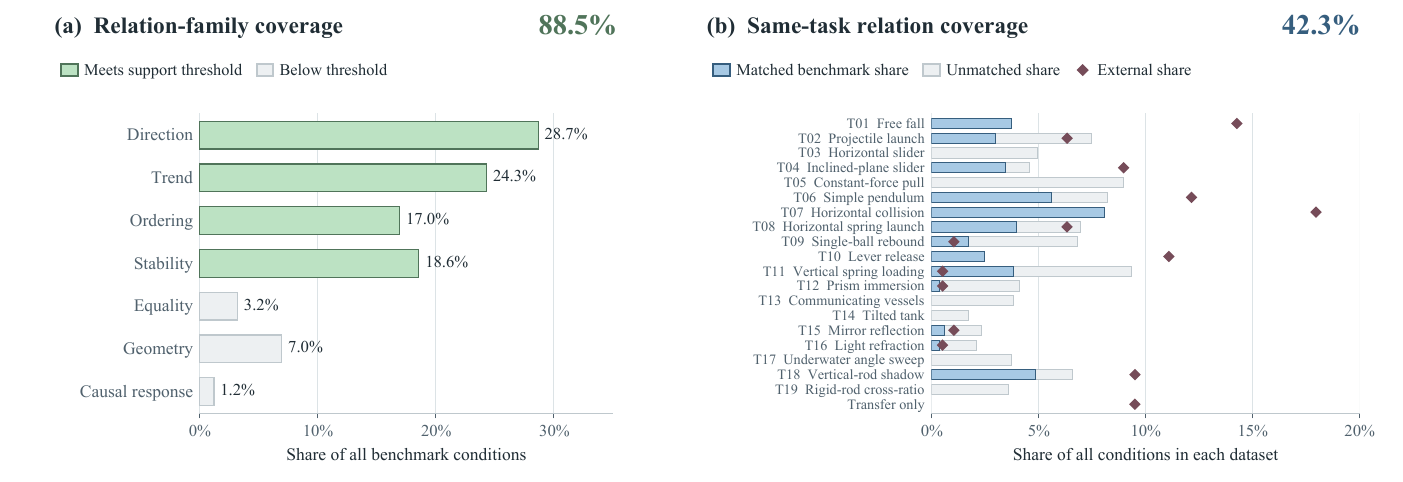}
\caption{Coverage of external $Q$ conditions. Top, shares are normalized within each dataset. Bottom, family coverage requires five independent videos and task--relation coverage requires one matched source. Both rates use the 802 registered visual conditions.}
\label{fig:q-coverage}
\end{figure*}

The PhyGround selection uses the task and configuration labels of the catalog frozen for that study (Figs.~\ref{fig:ei-task-coverage}(c) and~\ref{fig:ei-matches}). For $Q$, we compare external and benchmark conditions both by relation family and by the combination of task and relation (Fig.~\ref{fig:q-coverage}).

\subsection{Frozen evaluation prompts}
\label{app:evaluation-prompts}
The following boxes reproduce the frozen evaluation templates used by the benchmark. Braced fields are filled from the registered scene, objects, time window and evidence-frame list. The language is shown in English for readability; the JSON fields and admissibility checks are unchanged. The prompts do not expose a reference video or the expected numerical answer to the evaluator.

\begin{figure*}[!tp]
\centering
\begin{tcolorbox}[
  colback=promptblue,
  colframe=promptblue!55!black,
  coltitle=white,
  title={\textbf{L0 prompt: qualitative process realization (E)}},
  fonttitle=\small,
  fontupper=\small,
  width=\textwidth,
  boxrule=0.8pt,
  arc=3pt,
]
\noindent\textbf{System:} You are a strict video evaluation model.

\medskip
Evaluate physical realism for one physical law: \texttt{\{law\}}.

\smallskip
\noindent\textbf{Criterion:} \texttt{\{criteria\}}\\
\textbf{Caption, for context only:} \texttt{\{prompt\}}\\
\textbf{Sub-questions to consider before scoring:} \texttt{\{questions\_block\}}

\medskip
Judge the video itself. Do not penalize prompt mismatch unless it affects whether this physical law can be evaluated.

\medskip
\noindent\textbf{Score 1--5:}\\
5 = clearly correct; 4 = mostly correct with minor issues; 3 = partially correct or ambiguous; 2 = mostly incorrect; 1 = severely incorrect.

\medskip
Output only a JSON object with exactly one key: \texttt{\{law\}}.\\
\textbf{Example:} \texttt{\{"\{law\}": 3\}}
\end{tcolorbox}
\caption{Native 1--5 E prompt used for L0. Ratings are averaged over the applicable physical laws.}
\label{fig:prompt-l0}
\end{figure*}

\begin{figure*}[!p]
\centering
\begin{tcolorbox}[
  colback=promptcyan,
  colframe=promptcyan!55!black,
  coltitle=white,
  title={\textbf{I prompt: object integrity and persistence}},
  fonttitle=\small,
  fontupper=\small,
  width=\textwidth,
  boxrule=0.8pt,
  arc=3pt,
]
\noindent\textbf{System:} You are a strict video evaluation model.

\medskip
Evaluate Object Persistence.

\smallskip
\noindent\textbf{Caption, for context only:} \texttt{\{prompt\}}

\medskip
The video was generated using a text+image-to-video model, conditioned on the first frame and the text prompt above.

\smallskip
\noindent\textbf{Sub-questions to consider before scoring:} \texttt{\{questions\_block\}}

\medskip
\noindent\textbf{Score 1--5:}\\
5 = fully consistent; 4 = mostly consistent with minor flicker; 3 = noticeable issues; 2 = major inconsistencies; 1 = severe disappearance or identity changes.

\medskip
Output only a JSON object with exactly one key: \texttt{persistence}.\\
\textbf{Example:} \texttt{\{"persistence": 4\}}
\end{tcolorbox}
\caption{Native 1--5 I prompt for object persistence, evaluated separately from E and the L3--L4 measurements.}
\label{fig:prompt-integrity}
\par\smallskip
\begin{tcolorbox}[
  colback=promptyellow,
  colframe=promptyellow!55!black,
  coltitle=black,
  title={\textbf{L1/L2 prompt: binary qualitative relations (Q)}},
  fonttitle=\small,
  fontupper=\small,
  width=\textwidth,
  boxrule=0.8pt,
  arc=3pt,
]
You are a video physical qualitative-relation evaluator. Using only the reference frame, the ordered RGB frames, and the declared observation interval, judge every registered condition independently. Choose exactly one of \texttt{met} or \texttt{violated}: \texttt{met} means that the visible evidence overall supports the condition, and \texttt{violated} means that it supports the opposite. When evidence is ambiguous, choose the better-supported judgment and state the limitation.

\medskip
Compare only visible trajectories, relative positions, event order, trends, or geometric cues for the specified objects and window. Do not use the generation instruction, nominal parameters, a reference video, or what common sense predicts as evidence. Do not score appearance, object integrity, or an exact numerical ratio. Do not invent occluded frames, depth, speed, or events. Every judgment must contain at least one supplied frame index and a visible observation.

\medskip
\noindent\textbf{Output:} \texttt{\{input\_id, conditions\}}. Each condition appears exactly once with \texttt{condition\_id}, \texttt{judgment}, \texttt{evidence}, and \texttt{reason}. The score is computed afterward as \(Q=\texttt{met}/\texttt{(fixed registered conditions)}\). L1 uses an explicitly registered pair of videos; L2 uses the registered within-clip comparisons.
\end{tcolorbox}
\caption{Frozen Q prompt used for L1 intervention pairs and L2 within-clip relations. The condition list, objects, window, and evidence requirements are registered before inference.}
\label{fig:prompt-q}
\par\smallskip
\begin{tcolorbox}[
  colback=promptorange,
  colframe=promptorange!55!black,
  coltitle=black,
  title={\textbf{L3/L4 prompt: event and measurement setup}},
  fonttitle=\small,
  fontupper=\small,
  width=\textwidth,
  boxrule=0.8pt,
  arc=3pt,
]
First localize observable event boundaries from the labeled RGB frames. Return the registered event keys and original frame indices in the prescribed order; use \texttt{not\_reached} when a visible target has not reached a boundary and \texttt{null} when it cannot be localized. Define a steady interval from observed position or readout changes, never from agreement with theory. This pass only locates events and supplies visual evidence; it does not score the clip.

\medskip
Next configure image measurements from visible geometry. Return \texttt{geometry}, \texttt{features}, \texttt{measurement\_planes}, and \texttt{channels}. Use original pixel coordinates, no ruler, world scale, or expected outcome. Identify the actual coplanar apparatus surface for each displacement, liquid level, ray, pointer, or shadow; avoid background, transparent surfaces, and unsupported rays. For moving objects, provide a first-frame target description for tracking. For gravity, provide an object center for each ball; for optical scenes, define the interface normal; for instruments, read only visible dial values. If a channel cannot be independently measured, mark it unavailable and state why.

\medskip
The evaluator must not output a theoretical answer, fitted law, target score, or invented time series. After this prompt stage, deterministic tracking, depth estimation and law-specific code compute the dimensionless ratios and L3/L4 scores; when depth changes affect measurement, the registered correction is applied and recorded with the multiplier.
\end{tcolorbox}
\caption{Frozen L3/L4 acquisition prompt. It separates visual event and geometry proposals from the subsequent deterministic tracking, depth correction, physical-law calculation, and scoring.}
\label{fig:prompt-l34}
\end{figure*}

\section{Optical Measurement Validation}
\label{sec:historical}
\label{app:optical-validation}
We evaluate optical measurements on MP4-decoded simulated refraction clips with known ray geometry. The extractor measures the two ray directions independently from RGB without enforcing Snell's law. We calibrate nominal 95\% intervals on 39 videos and test on 40 independent videos. Table~\ref{tab:optical-validation} reports mean absolute error (MAE) across 320 test frames and interval coverage at one preselected frame per test video. The refracted-angle error is only $0.101^\circ$, yet its interval covers 85\% of test anchors, below the 95\% target.

\begin{table}[!ht]
\centering\small
\setlength{\tabcolsep}{3pt}
\caption{Optical measurement errors and coverage of nominal 95\% intervals. Angles are in degrees; the sine ratio is dimensionless.}
\label{tab:optical-validation}
\begin{tabular*}{\columnwidth}{@{\extracolsep{\fill}}lrrr@{}}
\toprule
Metric & MAE & Half-width & Coverage \\
\midrule
Refracted angle & 0.101 & 0.199 & 34/40 (85\%) \\
Snell angular residual & 0.050 & 0.106 & 39/40 (97.5\%) \\
Sine ratio & 0.00230 & 0.00615 & 40/40 (100\%) \\
\bottomrule
\end{tabular*}
\end{table}

\section{Generator Experiment Settings}
\label{app:generator-settings}

\paragraph{Benchmark scope.}
The benchmark contains 594 inputs across 23 tasks in six physics families and supports 1,231 evaluation samples (\cref{fig:inventory}). Their distribution is given in Appendix~\ref{app:input-record}.

\paragraph{Open-weight models.}
Pyramid Flow (SD3), Cosmos-Predict1-14B, Cosmos-Predict2-14B, Cosmos-Predict2.5-14B, Wan2.2-I2V-A14B and Cosmos3-Super-Image2Video use the same 594 benchmark inputs and 1,231 evaluation samples. We request one generated video per model and input with seed 42. Pyramid Flow (SD3) uses $1280\times768$ frames at 24 fps and 10 inference steps at each of its three stages. Cosmos-Predict1-14B-Video2World uses $1280\times704$ frames at 24 fps with 35 steps. Cosmos-Predict2-14B-Video2World and Cosmos-Predict2.5-14B use the same resolution and step count at 16 fps. We preserve the aspect ratio of the conditioning frame and pad it to the model canvas. Requested durations are 5, 8, 10 or 19 seconds. We round the frame count up to a length supported by the model. Pyramid Flow retains its default prompt suffix. For Predict1 requests exceeding 121 frames, we condition each continuation on the preceding segment's final frame with the same prompt and record the join positions. We store these outputs with lossless FFV1 encoding and verify the decoded frames.

\paragraph{API models.}
MiniMax-H3-Highspeed and Seedance 2.5 (\texttt{doubao-seedance-2-5-260628}) use the same 594 benchmark inputs and 1,231 evaluation samples, with both members of each L1 pair evaluated together. For each input the two services receive identical first-frame image bytes, prompt text and requested duration, with 480p output and adaptive aspect ratio. We retain the original provider files and use their decoded frame timestamps for measurements.

\paragraph{Qualitative scoring.}
Gemini 3.8 Flash applies the native PhyGround rubric~\cite{phyground} to the submitted prompt and temporally ordered video frames sampled at 4~Hz using the nearest original presentation time. It assigns one whole-video persistence rating $I$ on the native 1--5 scale. For $E$, it assigns a rating on the same scale to each registered applicable law and averages these ratings within the clip. At L0 we use this value only for eligible clips with a matching native law. For L1 and L2, GPT-6.1 Sol judges the frozen conditions against the ordered evidence for the designated pair or single clip. We divide the number of conditions met by the number registered in advance. A pair contributes once, and an absent partner or incomplete judgment remains unresolved. We average $I$ over scored videos and $E$ over eligible, scored L0 videos. For L1 and L2 we retain the number scored and the eligible denominator for each level.

\paragraph{Physical scoring and aggregation.}
We decode each original output directly and measure the registered objects over the prescribed event window. Depending on the task, the evaluator tracks object centers, reads surface heights or locates event times. When depth changes affect measurement, we detect and correct them (\cref{fig:inventory,fig:projection}). At L3 and L4, $P$ measures agreement with the law for one measured unit (Section~\ref{sec:method}). Coverage is the fraction of planned physical comparisons that could be measured. We retain the measurement and its residual with the score. For joint scoring, comparison weights are fixed within a sample, samples are weighted equally within a task, and tasks are weighted equally at each level. Missing evidence retains its weight in the lower and upper bounds (0--100). Whole-video ordinal $E$ and $I$ have no registered conversion to the unit-level gates. We keep their original scales and report the identifiable JPF interval.


\onecolumn
\section{Additional quantitative relations by scene}
\label{app:relation-catalog}
Table~\ref{tab:relation-catalog} gives eleven relations in addition to the five in Table~\ref{tab:relation-examples}. A scene appears in two rows when different stages of the experiment test different laws.

\noindent\begin{minipage}{\textwidth}
\centering
\captionsetup{hypcap=false}
\captionof{table}{Additional L3 and L4 tests, with their measurement windows and physical conditions. A dash means that the listed relation has no separate L4 test.}
\label{tab:relation-catalog}
\begingroup
\small
\setlength{\tabcolsep}{4pt}
\renewcommand{\arraystretch}{1.13}
\begin{tabularx}{\linewidth}{@{}>{\raggedright\arraybackslash}p{0.175\linewidth} >{\raggedright\arraybackslash}X >{\raggedright\arraybackslash}X@{}}
\toprule
Relation and task & L3: within-instance relation & L4: between-instance relation \\
\midrule
Constant acceleration\newline T04 incline; T05 constant pull &
From rest, along-track displacement obeys $s_2/s_1=(t_2/t_1)^2$. &
Over the same window of duration $\Delta t$,
$\frac{\Delta s_i}{\Delta s_j}=\frac{a_i(\tau_i+\Delta t/2)}{a_j(\tau_j+\Delta t/2)}$.
Here $\tau_i$ is time in motion before the window; release times may differ. T04: $a=g(\sin\alpha-\mu\cos\alpha)$. T05: $a=F_{\rm net}/m$.\\[2pt]
Uniform motion and range\newline T02, T03, T08 &
$\Delta x_2/\Delta x_1=\Delta t_2/\Delta t_1$. Use horizontal displacement for the projectile (T02), slider travel for T03, and coasting after spring separation for T08. &
T02: $R_i/R_j=1$ in the equal-range setup. T03: $\Delta x_i/\Delta x_j=v_i/v_j$ for equal time windows. T08: compare separation speeds as in the next row.\\[2pt]
Energy conversion\newline T08 spring launch; T09 inclined track &
T09: $v_2^2/v_1^2=\Delta h_2/\Delta h_1$ at two positions on one track, with both drops measured from release at rest. &
T08: $v_i/v_j=(x_i/x_j)\sqrt{k_i m_j/(k_j m_i)}$, where $x$ is initial spring compression. T09: $v_i^2/v_j^2=\Delta h_i/\Delta h_j$ across tracks.\\[2pt]
Collision and restitution\newline T07 blocks; T09 rebound &
T07: $v'_1/u_1=(m_1-e m_2)/(m_1+m_2)$ with block 2 initially at rest. T09: $h_{\rm rebound}/h_{\rm drop}=e^2$; measure both heights from the contact surface. &
T07: $(m_1-e m_2)v'_2=(1+e)m_1v'_1$. The post-collision velocities are signed; the test also holds when $v'_1=0$.\\[2pt]
Lever balance and rotation\newline T10 lever &
At equilibrium, $m_i r_i=m_j r_j$, where $r$ is the distance from pivot to attachment point. &
For the same angular rotation, the two attachment points satisfy $|\Delta y_i|/|\Delta y_j|=r_i/r_j$.\\[2pt]
Linear response\newline T11 spring; T22 compass; T23 bimetal strip &
Across two input stages, $Y_2/Y_1=X_2/X_1$. Use $(Y,X)=(\Delta\ell,F/k)$ for T11, $(\tan(\delta\theta),I)$ for T22, and $(\Delta\kappa,\Delta T)$ for T23. &
$Y_i/Y_j=X_i/X_j$ across matched objects or conditions. Measure spring extension from rest length, compass deflection from its initial direction, and curvature change from the initial strip shape.\\[2pt]
Floating depth\newline T12 prisms &
For a uniform straight prism at floating equilibrium, the submerged fraction is $f=h_{\rm sub}/H=\rho_{\rm obj}/\rho_{\rm liq}$. &
In one liquid, $f_i/f_j=\rho_i/\rho_j$. Both levels use the final frame and retain any deviation if the prisms have not settled.\\[2pt]
Free surface\newline T14 tilted tank &
In the final frame, $(y_R-y_L)/(x_R-x_L)=0$ between the two ends of the surface, relative to the physical horizontal. &
---\\[2pt]
Reflection, refraction and total internal reflection\newline T15--T17 interfaces &
Measure angles from the normal. Reflection: $\theta_r=\theta_i$. Refraction: $n_1\sin\theta_i=n_2\sin\theta_t$. If $n_1\sin\theta_i>n_2$, expect reflection and no transmission. &
For two rays in the same media, $\Delta\theta_r=\Delta\theta_i$ for reflection, and $\sin\theta_{t,1}/\sin\theta_{t,2}=\sin\theta_{i,1}/\sin\theta_{i,2}$ for refraction.\\[2pt]
Outgoing-ray agreement over time\newline T15--T17 ray paths &
Over frames $W$ in which the incident ray reaches the interface, $C=N_{\rm compliant}/N_W$. A missing required ray scores zero for that frame. Use the preceding row for total internal reflection. &
Compute $C$ separately for each designated path. A required outgoing ray absent throughout the window has $C=0$.\\[2pt]
Projective cross-ratio\newline T19 marked rod &
For four collinear marks, $C=\frac{\overline{AC}\,\overline{BD}}{\overline{AD}\,\overline{BC}}$ and $C_{\rm final}/C_{\rm initial}=1$. Use directed distances along the same axis. &
---\\
\bottomrule
\end{tabularx}
\endgroup
\end{minipage}

\par\smallskip
{\small\noindent\textit{Measurement conditions for Table~\ref{tab:relation-examples}.}
Free fall: measure drop $h$ and time $t$ from release at rest in the initial frame to each observation; use each ball's last descending state for L4, even if contact is unseen. Pendulum: compare consecutive full periods at small angles; measure $\ell$ from pivot to bob center in the first frame. Communicating vessels: $A$ is tube area; $h^0,h^f$ share a height datum. The change ratio applies to two tubes; each connected group also satisfies $\max_k h_k^f-\min_k h_k^f=0$ at equilibrium. Rod shadows: L3 uses a point lamp, L4 parallel light. $s,d$ are shadow length and rod-base distance from the lamp's floor projection; $H_i,H_L$ are rod and lamp heights. A fixed setup gives $\max_{t\in W}s_i(t)-\min_{t\in W}s_i(t)=0$ while lit. Magnet and coil: $J=\int_W[U(t)-U_0]dt$, $\bar U=J/\Delta t$. L3 compares opposite motions along one path; L4 compares same-direction insertions with equal flux changes and requires $J_iJ_j>0$.\par}

\clearpage
\section{Selected Evaluation Examples}
\label{app:sample-l0-l4}
Figures~\ref{fig:app-sample-l0-l4-minimax} and~\ref{fig:app-sample-l0-l4-seedance} show successful measurements on generated videos. Each row samples six frames uniformly across the full clip. Overlays show the recorded object-center tracks or liquid-surface positions. Scores are shown for the evaluated levels; L0 retains its native 1--5 scale.

\begin{center}
\begin{minipage}{\textwidth}
\captionsetup{hypcap=false}
\centering
\includegraphics[width=\textwidth]{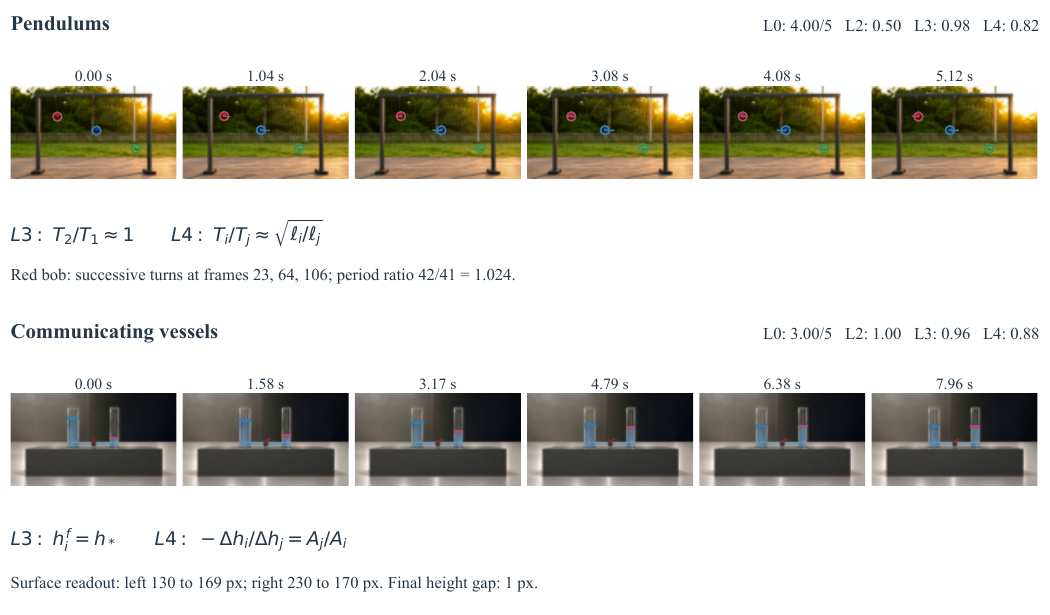}
\captionof{figure}{Successful measurement examples from MiniMax-H3: pendulum tracking and liquid-level readout. Recorded observations support the L3 and L4 relations shown beneath each sequence.}
\label{fig:app-sample-l0-l4-minimax}
\end{minipage}
\end{center}

\begin{center}
\begin{minipage}{\textwidth}
\captionsetup{hypcap=false}
\centering
\includegraphics[width=\textwidth]{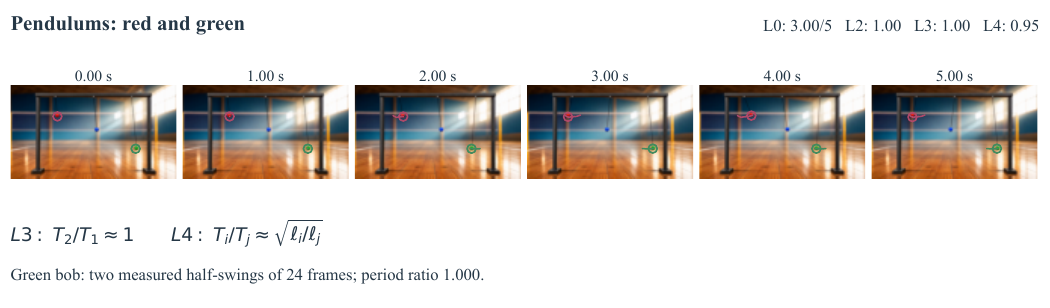}
\captionof{figure}{Successful measurement of the red and green pendulums from Seedance 2.5. L3 and L4 report the highlighted targets; L0 and L2 are clip-level scores.}
\label{fig:app-sample-l0-l4-seedance}
\end{minipage}
\end{center}

\clearpage
\section{Evaluator Validation Example}
\label{app:evaluator-validation-examples}
Figure~\ref{fig:app-evaluator-tracking-examples} shows successful two-target tracking on a public physics video. The evaluator receives RGB frames and target definitions and returns image-plane coordinates. Independent point annotations give median errors of 1.10 and 0.49 pixels for the two targets; all evaluated points are within five pixels.

\begin{center}
\begin{minipage}{\textwidth}
\captionsetup{hypcap=false}
\centering
\includegraphics[width=\textwidth]{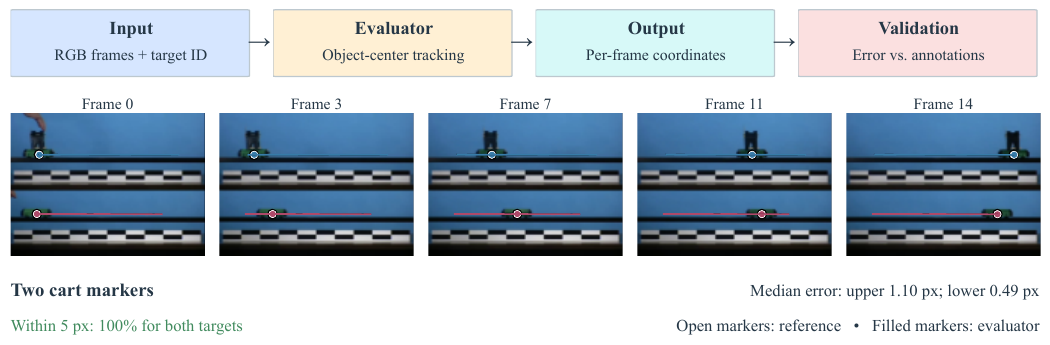}
\captionof{figure}{Successful two-target tracking. Open markers denote reference annotations and filled markers denote evaluator outputs. The five frames span the clip.}
\label{fig:app-evaluator-tracking-examples}
\end{minipage}
\end{center}

\end{document}